%% file: main.tex
\documentclass[10pt,twocolumn,letterpaper]{article}

\usepackage{wacv}
\usepackage{times}
\usepackage{epsfig}
\usepackage{graphicx}
\usepackage{amsmath}
\usepackage{amssymb}
\usepackage{times}
\usepackage{epsfig}
\usepackage{graphicx}
\usepackage{amsmath}

\usepackage{amssymb}
\usepackage{comment}
\usepackage{graphicx}
\usepackage{amsmath}
\usepackage{amssymb}
\usepackage{booktabs}
\usepackage{comment}
\usepackage{csquotes}
\usepackage{amsmath}
\usepackage{amssymb}
\usepackage{xcolor}
\usepackage{dsfont}
\usepackage{lipsum}
\usepackage{subcaption}
\usepackage{MnSymbol} 
\usepackage{amsthm}
\usepackage{enumitem}

\usepackage{algorithm}
\usepackage{algpseudocode}

\usepackage{stmaryrd}

\def\wacvPaperID{364} 

\wacvfinalcopy 

\ifwacvfinal
\usepackage[breaklinks=true,bookmarks=false]{hyperref}
\else
\usepackage[pagebackref=true,breaklinks=true,colorlinks,bookmarks=false]{hyperref}
\fi

\begin{document}

\title{A Visual Active Search Framework for Geospatial Exploration}

\author{Anindya Sarkar$^{1}$
\and
Michael Lanier$^{1}$
\and
Scott Alfeld$^{2}$
\and
Jiarui Feng$^{1}$
\and
Roman Garnett$^{1}$
\and
Nathan Jacobs$^{1}$~~~~~~~~Yevgeniy Vorobeychik$^{1}$\\
${}^1$\texttt{\{anindya,lanier.m,feng.jiarui,garnett,jacobsn,yvorobeychik\}@wustl.edu,}\\
${}^2$\texttt{salfeld@amherst.edu}\\${}^1$Department of Computer Science \& Engineering, Washington University in St. Louis\\
${}^2$Department of Computer Science, Amherst College\\}


\maketitle

\input{abstract}

\input{intro}

\input{relwork}
\input{model}

\input{method}
\vspace{-3pt}
\input{results}

\vspace{-4pt}
\section{Conclusion}
\vspace{-4pt}
Our results show that \emph{VAS} is an effective framework for geospatial broad area search. Notably, by applying simple \emph{TTA} techniques, the performance of \emph{VAS} can be further improved at test time in a way that is robust to target class shift. 
The proposed \emph{VAS} framework also suggests a myriad of future directions.
For example, it may be useful to develop more effective approaches for learning to search \emph{within a task}, as is common in past active search work.
Additionally, the search process may often involve additional constraints, such as constraints on the sequence of regions to query.
Moreover, it's natural to generalize query outcomes to be non-binary (e.g., returning the number of target object instances in a region).

\smallskip
\noindent\textbf{Acknowledgments }
This research was partially supported by the NSF (IIS-1905558, IIS-1903207, and IIS-2214141), ARO (W911NF-18-1-0208), Amazon, NVIDIA, and the Taylor Geospatial Institute.

{\small
\bibliographystyle{ieee_fullname}
\bibliography{egbib}
}

\input{supple}

\end{document}

%% file: abstract.tex

\begin{abstract}
Many problems can be viewed as forms of geospatial search aided by aerial imagery, with examples ranging from detecting poaching activity to human trafficking. %
We model this class of problems in a visual active search (VAS) framework, which has three key inputs: (1) an image of the entire search area, which is subdivided into regions, (2) a local search function, which determines whether a previously unseen object class is present in a given region, and (3) a fixed search budget, which limits the number of times the local search function can be evaluated. The goal is to maximize the number of objects found within the search budget.
We propose a reinforcement learning approach for VAS that learns a meta-search policy from a collection of fully annotated search tasks. This meta-search policy is then used to dynamically search for a novel target-object class, leveraging the outcome of any previous queries to determine where to query next.
%
Through extensive experiments on several large-scale satellite imagery datasets, we show that the proposed approach significantly outperforms several strong baselines. 
We also propose novel domain adaptation techniques that improve the policy at decision time when there is a significant domain gap with the training data.
Code is publicly available at this \href{https://github.com/anindyasarkarIITH/VAS}{$\>$\textcolor{blue}{link}}.
\end{abstract}

%% file: intro.tex
\vspace{-14pt}
\section{Introduction}
\vspace{-2pt}
  Consider a large national park that hosts endangered animals, which are also in high demand on a black market, creating a major poaching problem.
  An important strategy in an anti-poaching portfolio is to obtain aerial imagery using drones that helps detect poaching activity, either ongoing, or in the form of traps laid on the ground~\cite{bondi2018airsim,bondi2018spot,bondi2020birdsai,fang2015security,fang2016deploying}.
  The quality of the resulting photographs, however, is generally somewhat poor, making the detection problem extremely difficult.
  Moreover, park rangers can only inspect relatively few small regions to confirm poaching activity, doing so sequentially.
  Crucially, inspecting such regions yields \emph{new ground truth} information about poaching activity that we can use to decide which regions to inspect in the future.
  \begin{figure}[ht!]
  \centering
  \includegraphics[width=.8\linewidth]{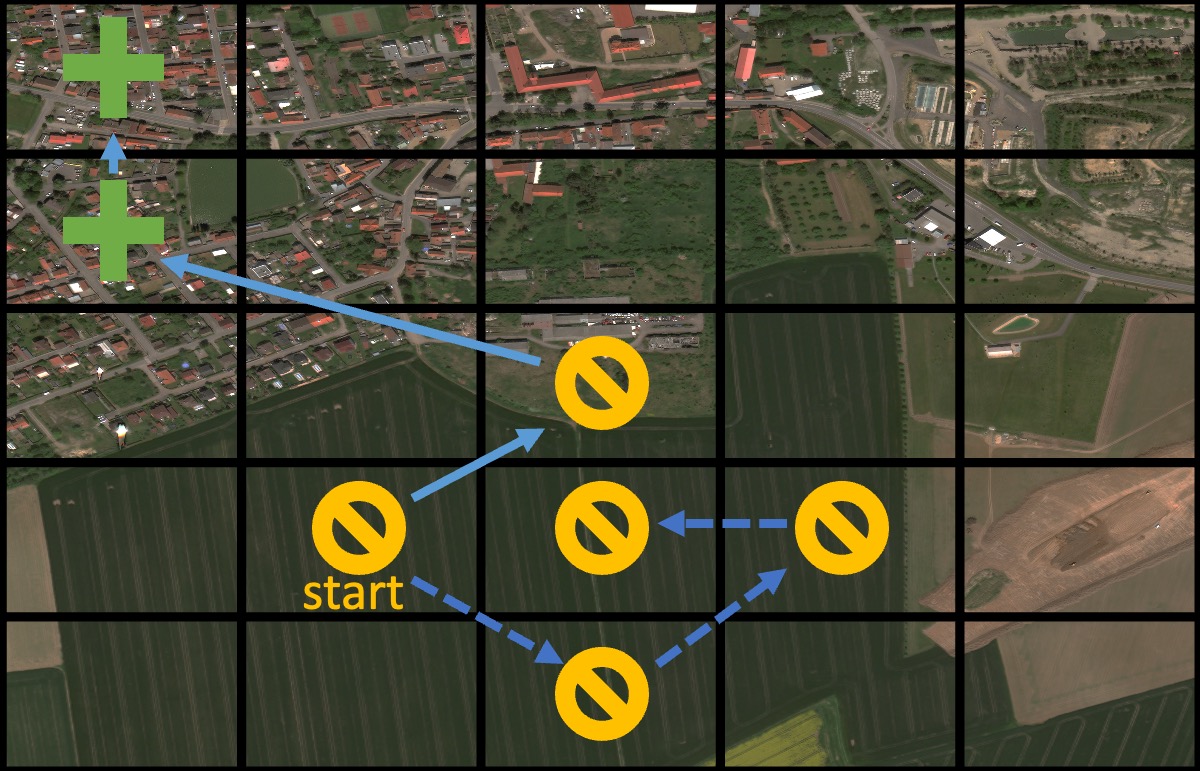}
 
  \caption{\small{A comparison of a greedy search policy (dashed line) with an active search strategy (solid line) for the \emph{small car} target class. The greedy policy is not able to adapt when a car is not found in the starting cell and needlessly searches many similar cells. The active strategy adapts and explores regions with different visual characteristics, eventually finding the objects of interest.
  }}
 
 \vspace{-10pt}
 \label{fig:introduction}
\end{figure}
  We can distill some key generalizable structure from this scenario: given a broad area image (often with a relatively low resolution), \emph{sequentially query} small areas within it (e.g., by sending park rangers to the associated regions, on the ground), with each query returning the ground truth, to identify as many target objects (e.g., poachers or traps) as possible.
  The number of queries we can make is typically limited, for example, by budget or resource constraints.
  Moreover, query results (e.g., detected poaching activity in a particular region) are \emph{highly informative about the locations of target objects in other regions}, for example, due to spatial correlation.
  We refer to this general modeling framework as \emph{visual active search (VAS)}.
  Numerous other scenarios share this broad structure, such as identification of drug or human trafficking sites, broad area search-and-rescue, identifying landmarks, and many others.

 
 

A simple solution to the broad-area search problem is to divide the broad area into many grid cells, train a classifier to predict existence of a target object in each grid cell, 
and simply explore the top $K$ cells in terms of predicted likelihood of the object being in the cell.
We call this the \emph{greedy} policy, which essentially reduces geospatial active search to the familiar \emph{object identification (or detection)} problem.
In Figure~\ref{fig:introduction}, we offer some intuition about why this idea fails to capture important problem structure.
Suppose we look for small cars in an image, starting in the grid marked \emph{start}, which we initially think is the best place to look.
The greedy policy being a one-shot predictor of likely grid cells containing target, continues to explore similar regions (marked as $\varobslash$).
What this approach ignores, as does framing the problem as traditional one-shot object identification, is the fact that both success and failure of past queries are informative due to complex spatial correlation among objects and other patterns in the scene; here, because the car was not found, we proceed to instead explore regions that have somewhat different characteristics.
%
%
The key to visual active search, therefore, is to learn how to make use of the ground truth information obtained over a sequence of queries to decide where to query next. Additionally, Section 5 of~\cite{garnett2012bayesian} provides a rigorous analysis of the general sub-optimality of greedy policies in active search settings.


\noindent
\textbf{Relationship to Active Search and Active Learning}
\emph{VAS} is closely related to \emph{active search}~\cite{garnett2012bayesian,garnett2015introducing,jiang2017efficient,jiang2019cost}.
Active search is typically concerned with binary classification, and aims to maximize the number of discovered positively-labeled inputs.
It uses a function $f$, which predicts labels of inputs $x$, as a means to this end, with each query serving the dual-purpose of improving $f$ as well as identifying a positive instance.
A central concern in active search, therefore, is achieving a balance of exploration (learning $f$) and exploitation (identifying target inputs).
This consideration is also the key distinction between active search and \emph{active learning}~\cite{settles2012active}, which is concerned \emph{solely} with improving the predictive quality of $f$.
Thus, if we had only a single query, active learning would typically choose $x$ for which prediction is highly uncertain, whereas active search would choose $x$ for which $f(x)$ is most confidently positive. We empirically show that active learning is inappropriate for solving the active search problem.

However, current active search approaches typically lack a pre-search training phase, and are therefore effective in relatively low dimensions and for relatively simple model classes such as $k$-nearest-neighbors.
In \emph{VAS}, in contrast, our goal is to \emph{learn how to search}, that is, to learn how to best use information obtained from previous search queries in choosing the next query.
We experimentally demonstrate the advantage of \emph{VAS} over conventional active search below.
 

\noindent
\textbf{Contributions}
We propose a deep reinforcement learning approach to solve the \emph{VAS} problem.
Our key contribution is a novel policy architecture which makes use of a natural representation of search state, in addition to the task image input, which the policy uses to dynamically adapt to the task at hand at decision time, without additional training.
Additionally, we consider a variant of \emph{VAS} in which the nature of input tasks at decision time is sufficiently different to warrant test-time adaptation, and propose several adaptation approaches that take advantage of the \emph{VAS} problem structure.
We extensively evaluate our proposed approaches to \emph{VAS} on two satellite imagery datasets in comparison with several baseline, including a state-of-the-art approach for a related problem of identifying regions of an image to zoom into~\cite{uzkent2020learning}.
Our results show that our approach significantly outperforms all baselines.

In summary, we make the following contributions:
\begin{itemize}[noitemsep,topsep=0pt]
    \item We propose \emph{visual active search (VAS)}, a novel visual reasoning model that represents an important class of geospatial exploration problems, such as identifying poaching activities, illegal trafficking, etc.
    \item We propose a deep reinforcement learning approach for \emph{VAS} that learns how to search for target objects in a broad geospatial area based on aerial imagery.
    \item We propose two new variants of \emph{test-time adaptation (TTA)} variants of \emph{VAS}: (a) \emph{Online TTA} and (b) \emph{Stepwise TTA}, as well as an improvement of the \emph{FixMatch} state-of-the-art \emph{TTA} method~\cite{sohn2020fixmatch}.
    \item We perform extensive experiments on two publicly available satellite imagery datasets, xView and DOTA, in a variety of settings, and demonstrate that proposed approaches significantly outperform all baselines.
\end{itemize}


%% file: relwork.tex
\section{Related Work}
\vspace{-4pt}
\noindent{\bf Foveated Processing of Large Images }
Numerous papers~\cite{wu2019liteeval, Uzkent_2020_WACV, uzkent2020learning, Yang_2020_CVPR, wang2020object, papadopoulos2021hard, thavamani2021fovea, meng2022adavit, meng2022count, Yang_2022_CVPR} have explored the use of low-resolution imagery to guide the selection of image regions to process at high resolution, including a number using reinforcement learning to this end. 
Our setting is quite different, as we aim to choose a sequence of regions to query, where each query yields the \emph{true label}, rather than a higher resolution image region, and these labels are important for both guiding further search, and as an end goal.

\smallskip
\noindent{\bf Reinforcement Learning for Visual Navigation }
Reinforcement learning has also been extensively used for visual navigation tasks, such as point and object localization~\cite{chaplot2020object,Mayo_2021_CVPR,Moghaddam_2022_WACV,Dwivedi_2022_CVPR}. 
While similar at the high level, these tasks involve learning to decide on a sequence of visual navigation steps based on a local view of the environment and a kinematic model of motion, and commonly do not involve search budget constraints.
In our case, in contrast, the full environment is observed initially (perhaps at low resolution), and we sequentially decide which regions to query, and are not limited to a particular kinematic model.

\smallskip
\noindent{\bf Active Search and Related Problem Settings }
Garnett et al.~\cite{garnett2012bayesian} first introduced \emph{Active Search (AS)}. Unlike \emph{Active Learning}~\cite{settles2009active}, \emph{AS} aims to discover members of valuable and rare classes rather than on learning an accurate model. 
Garnett et al.~\cite{garnett2012bayesian} 
demonstrated that for any $l > m$, a $l$-step lookahead policy can be arbitrarily superior than an $m$-step one, showing that a nonmyopic active search approach can be significantly better than a myopic one-step lookahead.
Jiang et al.~\cite{jiang2017efficient,jiang2018efficient} proposed approaches for efficient nonmyopic active search, while
Jiang et al.~\cite{jiang2019cost} introduced consideration of search cost into the problem.
We note two crucial differences between our setting and the previous works on active search.
First, we are the first to consider the problem in the context of vision, where the problem is high-dimensional, while prior techniques rely on a relatively low dimensional feature space.
Second, we use reinforcement learning as a means to learn a search policy, in contrast to prior work on active search which  aims to design efficient search algorithms.

%% file: model.tex
\section{Model}

At the center of our task is an aerial image $x$ which is partitioned into $N$ grid cells,
$x = (x^{(1)}, x^{(2)},..., x^{(N)})$. We can also view $x$ as the disjoint union of these $N$ grid cells, each of which is a sub-image. 
A subset (possibly empty) of these grid cells contain an instance of the target object.
We formalize this by associating each grid cell $j$ with a binary label $y^{(j)} \in \{0,1\}$, where $y^{(j)} = 1$ iff grid cell $j$ contains the target object. Let $y = (y^{(1)}, y^{(2)},...,y^{(N)})$.


We do not know $y$ a priori, but can sequentially query to identify grid cells that contain the target object. 
Whenever we query a grid cell $j$, we obtain both the associated label $y^{(j)}$, (i.e., whether it contains the target object) \emph{and} accrue utility if the queried cell actually contains a target object.
Our ultimate goal is to find as many target objects as possible through a sequence of such queries given a total query budget constraint $\mathcal{C}$, 

Formally, let $c(j,k)$ be the cost of querying grid cell $k$ if we start in grid cell $j$.
For the very first query, we can define a dummy initial grid cell $d$, so that cost function $c(d,k)$ captures the initial query cost.
Let $q_t$ denote a query performed in step $t$.
Our ultimate goal is to solve the following optimization problem:
\begin{equation}
\begin{split}
    &\max_{\{q_t\}} U(x;\{q_t\}) \equiv \sum_{t} y^{(q_t)}\\
    &\mathrm{s.t.:} \sum_{t\ge 0} c(q_{t-1},q_t) \le \mathcal{C},
\end{split}
\end{equation}
where $c(q_{-1},q_0) = c(d,q_0)$.


In order to succeed, we need to use the labels from previously queried cells to decide which cell to query next.
This is a conventional setup in active search, where an important means to accomplish such learning is by introducing a model $f$ to predict a mapping between (in our instance) a grid cell and the associated label (whether it contains a target object)~\cite{garnett2015introducing,jiang2017efficient,jiang2019cost}.
However, in many domains of interest, such as most visual domains of the kind we consider, the query budget $\mathcal{C}$ and the number of grid cells $N$ are very small compared to the dimension of the input $x$, far too small to learn a meaningful prediction $f$.
Instead, we suppose that we have a dataset of \emph{tasks} (aerial images) for which we have labeled whether each of the grid cells contains the target object.
Let this dataset be denoted by $\mathcal{D} = \{(x_i,y_i)\}$, with each $x_i = (x_i^{(1)},x_i^{(2)},\ldots,x_i^{(N)})$ the task image and $y_i = (y_i^{(1)},y_i^{(2)},\ldots,y_i^{(N)})$ its corresponding grid cell labels.
Then, at decision (or inference) time, we observe the task aerial image $x$, including its partition into the grid cells, and choose queries $\{q_t\}$ sequentially to maximize $U(x;\{q_t\})$.

We consider two variations of the model above.
In the first, each instance $(x_i,y_i)$ in the training data $\mathcal{D}$, as well as $(x,y)$ at decision time (when $y$ is unobserved before queries) are generated i.i.d.~from the same distribution.
In the second variation, while instances $(x,y)$ are still i.i.d.~at decision time, their distribution can be different from that of the training data $\mathcal{D}$.
The latter variation falls within the broader category of \emph{test-time adaptation (TTA)} settings, but with the special structure pertinent to our model above.

%% file: method.tex
\vspace{-3pt}
\section{Solution Approach}
\vspace{-3pt}

\begin{figure}
 \centering
 \includegraphics[height=1.2in,width=2.5in]{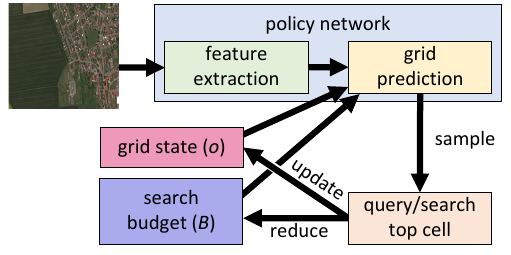}
 \caption{An overview of the \textit{VAS} framework.}
 \vspace{-10pt}
 \label{fig:vas_framework}
\end{figure}

\begin{figure*}
  \centering
    \includegraphics[width=.65\linewidth]{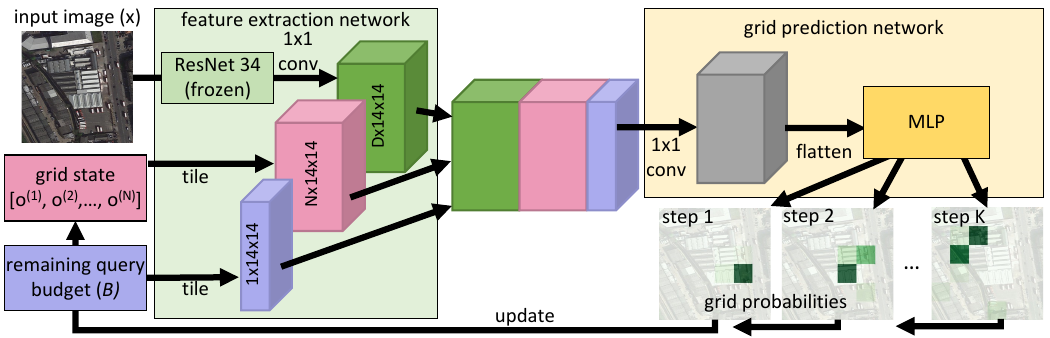}
  \caption{Our VAS policy network architecture, showing the grid probabilities at three different steps.}
  \label{fig:feat_ext}
\end{figure*}
Visual active search over the area defined by $x$ and its constituent grid cells is a dynamic decision problem.
As such, we model it as a budget-constrained episodic Markov decision process (MDP), where the search budget $\mathcal{C}$ is defined for each instance $x$ \emph{at decision time}.
In this MDP, the actions are simply choices over which grid cell to query next; we denote the set of grids by $A = \{1,\ldots,N\}$.
Since in our model there is never any value to query a grid cell more than once, we restrict actions available at each step to be only grids that have not yet been queried (in principle, this restriction can also be learned).
Policy network inputs include: 1) the overall input $x$, which is crucial in providing the broad perspective on each search problem, 2) outcomes of past search queries $o$ (we detail our representation of this presently), and 3) remaining budget $B \le \mathcal{C}$.
State transition simply updates the remaining budget and adds the outcome of the latest search query to state.
Finally, an immediate reward for query a grid cell $j$ is $R(x,o,j) = y^{(j)}$.
We represent outcomes of search query history $o$ as follows.
Each element of $o$ corresponds to a grid cell $j$, so that $o = (o^{(1)},\ldots,o^{(N)})$.
$o^{(j)} = 0$ if $j$ has not been previously queried.
If grid cell $j$ has been previously queried, 
\begin{equation}
    o^{(j)}\xleftarrow{}\left\{
    \begin{array}
    {r@{\quad}l}
         \text{$1,$} & \text{if\>\>\>$y^{(j)} = 1$} \\
         \text{$-1,$}  & \text{if\>\>\>$y^{(j)} = 0$.} 
    \end{array}
    \right.
    \label{eq:observation}
\end{equation}

Armed with this MDP problem representation, we next describe our proposed deep reinforcement learning approach for learning a search policy that makes use of a dataset $\mathcal{D}$ of past search tasks.
Specifically, we use the \emph{REINFORCE} policy gradient algorithm~\cite{sutton1999policy} to directly learn a search policy $\psi(x,o,B;\theta)$ where $\theta$ are the parameters of the policy that we learn. Specifically, we maximize the following objective function:
\begin{equation}
\label{eq:REINFORCE}
\nabla J(\theta) = \sum_{i=1}^{M} \sum_{t=1}^{T_i} \mathds{1}_{\sum_{t \ge 0} c(q_{t-1},q_t) \le \mathcal{C}} \nabla \log \psi_{\theta}(a_t^i|x_i,o_t^i,B_t^i) \mathbb{R}_t^i
\end{equation}
Where $M$ is the number of example search task seen during training and $\mathbb{R}_t$ is the discounted cumulative reward defined as 
$\mathbb{R}_t = \sum_{k=t}^{T} \gamma^{k-t} R_k$ with a discount factor \(\gamma \in [0, 1]\).

The output of the search policy $\psi$ is a probability distribution over $A$, with $\psi_j(x,o,B;\theta)$ the probability that grid cell $j \in A$ is selected by the policy $\psi$.

In general, $\psi$ will output a positive probability over all possible grid cells $j \in A$.
However, in our setting there is no benefit to querying any grid cell $j \in A$ that has previously been queried, i.e., for which $o^{(j)} \ne 0$.
Consequently, both at training (when the next decision is generated) and decision time, we restrict consideration only to $j \in A$ with $o^{(j)} = 0$, that is, which have yet to be queried, and simply renormalize the output probabilities of $\psi$.
Formally, we define $\psi_j'(x,o,B;\theta) = 0$ for $j$ with $o^{(j)} \ne 0$, and define 
$$\psi_j'(x,o,B;\theta) = \frac{\psi_j(x,o,B;\theta)}{\sum_{k \in A : o^{(k)} = 0} \psi_k(x,o,B;\theta)}.$$
Grid cells $j$ are then samples from $\psi_j'$ at each search step during training.
At decision time, on the other hand, we choose the grid cell $j$ with the maximum value of $\psi_j'$.
This approach allows us to simply train the policy network $\psi$ without concern about feasibility of particular grid cell choices at decision time.
In addition, to ensure that the policy is robust to search budget uncertainty, we use randomly generated budgets $\mathcal{C}$ at training time for different task instances. 
In the case of query costs $c(j,k)=1$ for all grid cells $j,k$, each episode has a fixed length $\mathcal{C}$.
In general, episodes have no fixed length, and end whenever we exhaust the total cost budget $\mathcal{C}$.
The overview of our proposed \textit{VAS} framework is depicted in Figure~\ref{fig:vas_framework}.

Next, we detail the proposed policy network architecture, and subsequently describe an adaptation of our approach when instances at decision time follow a different distribution from those in the training data $\mathcal{D}$, that is, the test-time adaptation (TTA) setting.

\subsection{Policy Network Architecture}

As shown in Figure~\ref{fig:feat_ext}, the policy network $\psi(x,o,B;\theta)$ is composed of two components: 1) the image feature extraction component $f(x;\phi)$ which maps the aerial image $x$ to a low-dimensional latent feature representation $z$, and 2) the grid selection component $g(z,o,B;\zeta)$, which combines the latent image representation $z$ with outcome of past search queries $o$ and remaining budget $B$ to produce a probability distribution over grid cells to search in the next time step.
Thus, the joint parameters of $\psi$ are $\theta = (\phi, \zeta)$.

We use a frozen ResNet-34~\cite{he2016deep}, pretrained on ImageNet~\cite{krizhevsky2017imagenet}, as the feature extraction component $f$, followed by a $1\times 1$ convolution layer.
We combine this with the budget $B$ and past query information $o$ as follows.
We apply the tiling operation in order to convert $o$ into a representation with the same dimensions as the extracted features $z = f(x)$, 
aiding us to effectively combine latent image feature and auxiliary state feature while preserving the grid specific spatial and query related information.
Similarly, we apply tiling to the scalar budget $B$ to transform it to match the size of $z$ and the tiled version of $o$.
Finally, we concatenate the features $(z,o,B)$ along the channels dimension and pass them through the grid prediction network $g$. This consists of $1\times1$ convolution to reduce dimensionality, flattening, a small MLP with ReLU activations, and a final output (softmax) that represents the current grid probability. This yields the full policy network to be trained end to end via REINFORCE:
\(
\psi(x,o,B;\theta) = g(f(x;\phi),o,B;\zeta).
\)

\subsection{Test-Time Adaptation}

\label{S:tta}

A central issue in our model, as in traditional active search, is that tasks faced at decision time may in some respects be novel, unlike tasks faced previously (e.g., represented in the dataset $\mathcal{D}$).
We view this issue through the lens of \emph{test-time adaptation (TTA)}, in which predictions are made on data that comes from a different distribution from training data.
While a myriad of TTA techniques have been developed, they have focused almost exclusively on \emph{supervised learning} problems, rather than active search settings of the kind we study.
Nevertheless, two common techniques can be either directly applied, or adapted, to our setting: 1) \emph{Test-Time Training (TTT)}~\cite{sun2020test} and 2) \emph{FixMatch}~\cite{sohn2020fixmatch}.

\emph{TTT} makes use of a self-supervised objective at both training and prediction time by adding a self-supervised head $r$ as a component of the policy model.
The associated self-supervised loss (which is added during training) is a quadratic reconstruction loss $||x-r(z;\eta)||$, where $z = f(x;\phi)$ is the latent embedding of the input aerial image $x$ and $\eta$ the parameters of $r$.
At decision time, a new task image $x$ is used to update policy parameters using just the reconstruction loss before we begin the search.
Adaptation of \emph{TTT} to our VAS domain is therefore direct.

The original variant of \emph{FixMatch} uses pseudo-labels at decision time, which are predictions on weakly augmented variants of the input image $x$ (keeping only those which are highly confident), to update model parameters.
In our domain, however, we can leverage the fact that we obtain \emph{actual} labels whenever we query regions of the image.
We make use of this additional information as follows.
Whenever a query $j$ is successful (i.e., $y^{(j)}=1$), we construct a label vector as the one-hot vector with a 1 in the location of the successful grid cell $j$.
However if $y^{(j)}=0$, we associate each queried grid cell with a 0, and assign uniform probability distribution over all unqueried grids.
We then update model parameters using a cross-entropy loss.

Even as we adapted them, \emph{TTT} and \emph{FixMatch} do not fully take advantage of the rich information obtained at decision time in the \emph{VAS} context as we proceed through each input task: we not only observe the input image $x$, but also observe query results over time during the search.
We therefore propose two new variants of \emph{TTA} which are specific to the \emph{VAS} setting: (a) \emph{Online TTA} and (b) \emph{Stepwise TTA}.
In \emph{Online TTA}, we update parameters of the policy network after each task is completed during decision time, which yields for us both the input $x$ and the \emph{observations} $o$ of the search results, which only partially correspond to $y$, since we have only observed the contents of the previously queried grid cells.
Nevertheless, we can simply use this partial information $o$ as a part of the \emph{REINFORCE} policy gradient update step to update the policy parameters $\theta$.
In \emph{Stepwise TTA}, we update the policy network parameters, even during the execution of a particular task, at decision time, once every $m < \mathcal{C}$ steps.
The main difference between \emph{Online} and \emph{Stepwise} variations of our \emph{TTA} approaches is consequently the frequency of updates.
Note that we can readily compose both of these \emph{TTA} approaches with conventional \emph{TTA} methods, such as \emph{TTT} and \emph{FixMatch}.



%% file: results.tex
\section{Experiments}
\vspace{-4pt}
\noindent\textbf{Evaluation Metric }
We evaluate the proposed approaches in terms of the average number of target objects discovered (we shorten it to \textbf{ANT}).
\smallskip
\noindent\textbf{Baselines }
We compare the proposed VAS policy learning framework with the following baselines:
\begin{enumerate}[noitemsep,topsep=0pt]
\item \emph{random search}, where each grid is chosen uniformly at random among those which haven't been explored,
\item \emph{greedy classification}, in which we train a classifier $\psi_\mathit{gc}$ to predict whether a particular grid has a target object and search the grids most likely to contain the target until the search budget is exhausted, and
\item \emph{greedy selection}, based on the approach by Uzkent and Ermon~\cite{uzkent2020learning} which trains a policy $\psi_\mathit{gs}$ which yields a probability of zooming into each grid cell $j$.  We select grids according to $\psi_\mathit{gs}$ until the budget $\mathcal{C}$ is saturated.
\item \emph{active learning}, in which we randomly select the first grid to query and then choose $\mathcal{C}-1$ grids using a state-of-the-art active learning approach by Yoo et al.~\cite{yoo2019learning}.
\item \emph{conventional active search}, an active search method by Jiang et al.~\cite{jiang2017efficient}, using a low-dimensional feature representation for each image grid from the same feature extraction network as in our approach.
\end{enumerate}

\smallskip
\noindent\textbf{Query Costs} We consider two ways of generating query costs: (i) $c(i,j) = 1$ for all $i,j$, where $\mathcal{C}$ is just the number of queries, and (ii) $c(i,j)$ is based on Manhattan distance between $i$ and $j$.
Most of the results we present reflect the second setting; the results for uniform query costs are qualitatively similar and provided in the Supplement.

\smallskip
\noindent\textbf{Datasets } We evaluate the proposed approach using two datasets: xView~\cite{lam2018xview} and DOTA~\cite{xia2018dota}.
xView is a satellite imagery dataset which consists of large satellite images representing 60 categories, with approximately 3000 pixels in each dimensions. We use $67\%$ and $33\%$ of the large satellite images to train and test the policy network respectively. 
DOTA is also a satellite imagery dataset.
We re-scale the original $\sim$ 3000 $\times$ 3000px images to 1200 $\times$ 1200px.
Unless otherwise specified, we use $N=36$
non-overlapping pixel grids each of size 200 $\times$ 200. 

\subsection{Results on the xView Dataset}
We begin by evaluating the proposed approaches on the xView dataset, varying search budgets $\mathcal{C} \in \{25, 50, 75\}$ and number of grid cells $N \in \{30, 48, 99\}$.
We consider two target classes: \emph{small car} and \emph{building}.
As the dataset contains variable size images, take random crops of $2500 \times 3000$ for $N=30$, $2400 \times 3200$ pixels for $N=48$, and $2700 \times 3300$ for $N=99$, thereby ensuring equal grid cell sizes.

\begin{table}[h!]
       \scriptsize
        \centering        \caption{\footnotesize{\textbf{ANT} comparisons for the \textit{small car} target class on xView.}}
        \begin{tabular}{lcccc}
        \toprule
             \textbf{Method} & $\mathcal{C}=25$ & $\mathcal{C}=50$ & $\mathcal{C}=75$  \\
            \midrule
             \emph{random search} ($N=30$)  & 3.41 & 3.95 & 4.52 \\
             \emph{greedy classification} ($N=30$) & 3.91 & 4.60 & 4.76 \\
             \emph{greedy selection}~\cite{uzkent2020learning} ($N=30$)  & 3.90 & 4.63 & 4.78 \\
             \emph{active learning}~\cite{yoo2019learning} ($N=30$)  & 3.92 & 4.58 & 4.73 \\
             \emph{conventional active search}~\cite{jiang2017efficient} ($N=30$)  & 3.61 & 4.17 & 4.70 \\
             \midrule
             \textit{\textbf{VAS}} ($N=30$) & \textbf{4.61} & \textbf{7.49} & \textbf{9.88} \\
             \midrule
             \midrule
             \emph{random search} ($N=48$)  & 3.20 & 3.66 & 4.11 \\
             \emph{greedy classification} ($N=48$) & 3.87 & 4.29 & 4.52 \\
             \emph{greedy selection}~\cite{uzkent2020learning} ($N=48$)  & 3.89 & 4.42 & 4.53 \\
             \emph{active learning}~\cite{yoo2019learning} ($N=48$)  & 3.87 & 4.28 & 4.51 \\
             \emph{conventional active search}~\cite{jiang2017efficient} ($N=48$)  & 3.26 & 3.74 & 4.32 \\
             \midrule
             \textit{\textbf{VAS}} ($N=48$) & \textbf{4.56} & \textbf{7.45} & \textbf{9.63} \\
             \midrule
             \midrule
             \emph{random search} ($N=99$)  & 1.10 & 2.15 & 2.96 \\
             \emph{greedy classification} ($N=99$) & 1.72 & 2.79 & 3.36 \\
             \emph{greedy selection}~\cite{uzkent2020learning} ($N=99$)  & 1.78 & 2.83 & 3.41 \\
             \emph{active learning}~\cite{yoo2019learning} ($N=99$)  & 1.69 & 2.78 & 3.33 \\
             \emph{conventional active search}~\cite{jiang2017efficient} ($N=99$)  & 1.42 & 2.31 & 3.10 \\
             \midrule
             \textit{\textbf{VAS}} ($N=99$) & \textbf{2.72} & \textbf{4.42} & \textbf{5.78} \\
             \bottomrule
        \end{tabular}
        \vspace{-4pt}
        \label{tab:xview-smallcar}
\end{table}
\vspace{-4pt}
\begin{table}[h!]
       \scriptsize
        \centering        \caption{\footnotesize{\textbf{ANT} comparisons for the \textit{building} target class on xView.}}
        \begin{tabular}{lcccc}
        \toprule
             \textbf{Method} & $\mathcal{C}=25$ & $\mathcal{C}=50$ & $\mathcal{C}=75$  \\
            \midrule
             \emph{random search} ($N=30$)  & 3.97 & 4.94 & 5.39 \\
             \emph{greedy classification} ($N=30$) & 4.69 & 5.27 & 5.80 \\
             \emph{greedy selection}~\cite{uzkent2020learning} ($N=30$)  & 4.84 & 5.33 & 5.82 \\
             \emph{active learning}~\cite{yoo2019learning} ($N=30$)  & 4.67 & 5.24 & 5.80 \\
             \emph{conventional active search}~\cite{jiang2017efficient} ($N=30$)  & 4.15 & 5.20 & 5.51 \\
             \midrule
             \textit{\textbf{VAS}} ($N=30$) & \textbf{5.65} & \textbf{9.31} & \textbf{12.20} \\
             \midrule
             \midrule
             \emph{random search} ($N=48$)  & 3.47 & 3.96 & 4.26 \\
             \emph{greedy classification} ($N=48$) & 3.90 & 4.43 & 4.61 \\
             \emph{greedy selection}~\cite{uzkent2020learning} ($N=48$)  & 3.95 & 4.51 & 4.67 \\
             \emph{active learning}~\cite{yoo2019learning} ($N=48$)  & 3.88 & 4.43 & 4.60 \\
             \emph{conventional active search}~\cite{jiang2017efficient} ($N=48$)  & 3.70 & 4.11 & 4.38 \\
             \midrule
             \textit{\textbf{VAS}} ($N=48$) & \textbf{5.61} & \textbf{9.26} & \textbf{12.15} \\
             \midrule
             \midrule
             \emph{random search} ($N=99$)  & 1.55 & 2.99 & 4.18 \\
             \emph{greedy classification} ($N=99$) & 2.17 & 3.96 & 4.84 \\
             \emph{greedy selection}~\cite{uzkent2020learning} ($N=99$)  & 2.29 & 4.21 & 5.22 \\
             \emph{active learning}~\cite{yoo2019learning} ($N=99$)  & 2.17 & 3.95 & 4.82 \\
             \emph{conventional active search}~\cite{jiang2017efficient} ($N=99$)  & 1.68 & 3.10 & 4.33 \\
             \midrule
             \textit{\textbf{VAS}} ($N=99$) & \textbf{4.29} & \textbf{6.91} & \textbf{8.98} \\
             \bottomrule
        \end{tabular}
        \label{tab:xview-building}
\end{table}
The results are presented in Table~\ref{tab:xview-smallcar} for the \emph{small car} class and in Table~\ref{tab:xview-building} for the \emph{building} class.
We see substantial improvements in performance of the proposed \emph{VAS} approach compared to all baselines, ranging from $15$--$260\%$ improvement relative to the most competitive state-of-the-art approach, \emph{greedy selection}.
There are two general consistent trends.
First, as the number of grids $N$ increases compared to $\mathcal{C}$ (corresponding to sets of rows in either table), performance of all methods declines, as the task becomes more challenging.
However, the decline in performance is typically much greater for our baselines than for \emph{VAS}.
Second, overall performance improves as $\mathcal{C}$ increases (columns in both tables), and the relative advantage of \emph{VAS} increases, as it is better able to take advantage of the greater budget than the baselines.


\begin{figure}[h!]
    \centering
    \begin{subfigure}[b]{0.23\textwidth}
        \centering
    \includegraphics[width=.9\linewidth]{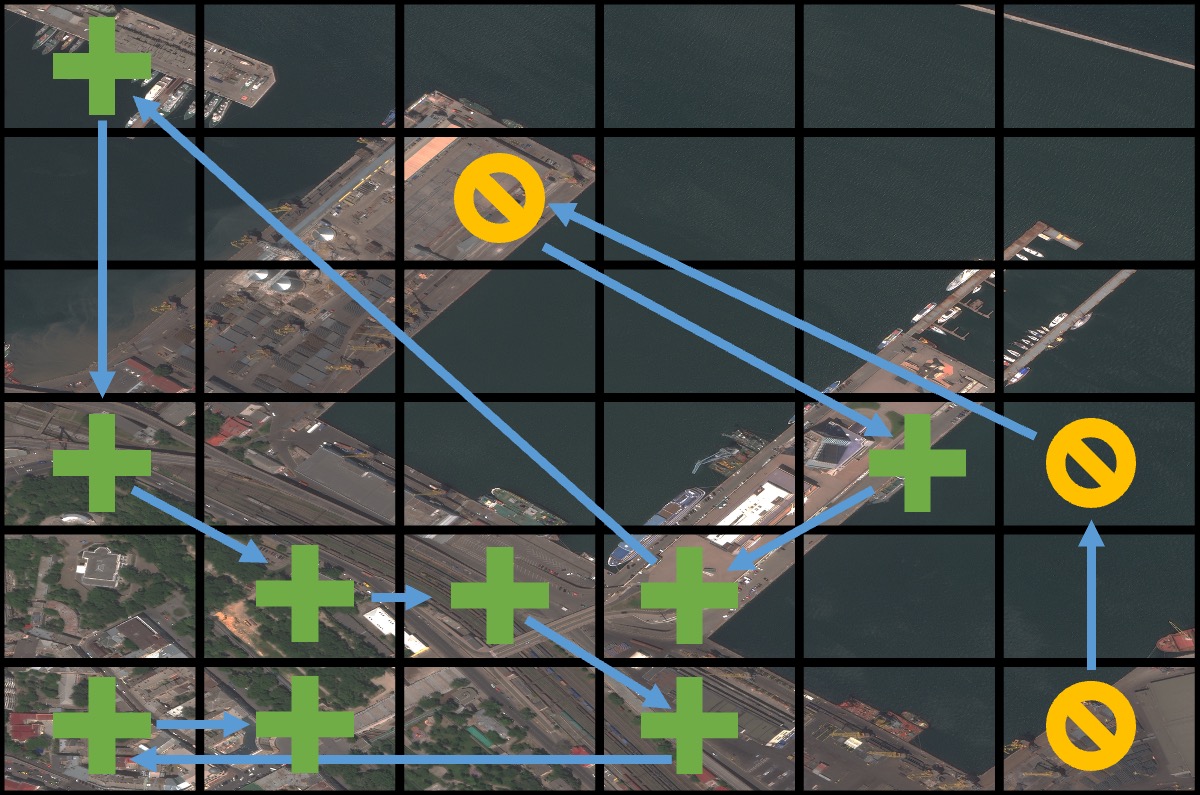}
        \label{fig:vis_vas}
    \end{subfigure}%
    \begin{subfigure}[b]{0.23\textwidth}
        \centering
    \includegraphics[width=.9\linewidth]{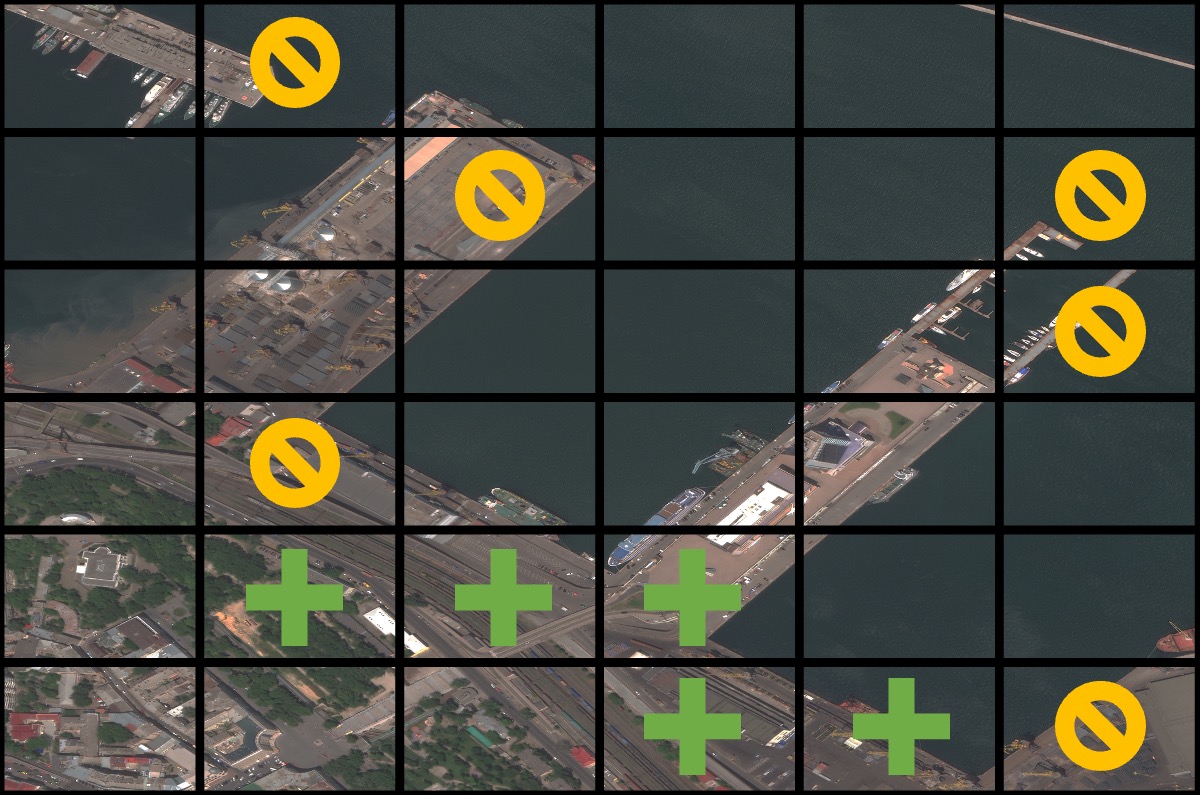}
        \label{fig:vis_greedy}
    \end{subfigure}
    \caption{\footnotesize{Comparison of policies learned using \textit{VAS} (left) and the \emph{greedy selection} baseline method (right).} 
    }
    \vspace{-2pt}
    \label{fig:vis_vas0}
\end{figure}
In Figure~\ref{fig:vis_vas0} we visually illustrate \emph{VAS} search strategy in comparison with the \emph{greedy selection} baseline (the best performing baseline).
The plus signs correspond to successful queries, $\varobslash$ to unsuccessful queries, and arrows represent query order.
This shows that \emph{VAS} quickly learns to take advantage of the visual similarities between grids (after the first several failed queries, the rest are successful), whereas our most competitive baseline---\emph{greedy selection}---fails to take advantage of such information. During the initial search phase, the VAS policy explores different types of grids before exploiting grids it believes to have target objects. 

Finally, we perform an ablation study to understand the added value of including remaining budget $B$ as an input in the \emph{VAS} policy network.
To this end, we modify the combined feature representation of size $(2N+1) \times 14 \times 14$, consisting of input and auxiliary state features, each of size $N \times 14 \times 14$, and a single channel of size $14 \times 14$ containing the information of remaining search budget, as depicted in Figure~\ref{fig:feat_ext}. We only eliminate the channel from the combined feature representation that contains the information about the number of queries left, resulting in $2N \times 14 \times 14$ size feature map. 
The resulting policy network is then trained just as the original \emph{VAS} architecture.

\begin{table}[h!]
       \scriptsize
        \centering
        \caption{\footnotesize{Comparative \textbf{ANT} performance of \textit{VAS without remaining search budget} and \textit{VAS} using \emph{small car} as the target class.}}
        \begin{tabular}{lcccc}
        \toprule
             \textbf{Method} &  $\mathcal{C}=25$ & $\mathcal{C}=50$ & $\mathcal{C}=75$ \\
            \midrule
             \textit{\textbf{VAS w/o remaining search budget}} ($N=30$)  & 4.47 & 7.38 & 9.62 \\
             \textit{\textbf{VAS}} ($N=30$) & \textbf{4.61} & \textbf{7.49} & \textbf{9.88} \\
             \midrule
             \textit{\textbf{VAS w/o remaining search budget}} ($N=48$)  & 4.34 & 7.31 & 9.49 \\
             \textit{\textbf{VAS}} ($N=48$) & \textbf{4.56} & \textbf{7.45} & \textbf{9.63} \\
             \midrule
             \textit{\textbf{VAS w/o remaining search budget}} ($N=99$)  & 2.63  & 4.29 & 5.69 \\
             \textit{\textbf{VAS}} ($N=99$)  & \textbf{2.72}  & \textbf{4.42} & \textbf{5.78} \\
             \bottomrule
        \vspace{-3pt}
        \end{tabular}
        \vspace{-3pt}
        \label{tab:ablationrsb}
\end{table}
\vspace{-3pt}
We compare the performance of the policy without remaining search budget (referred to as \emph{VAS without remaining search budget}) with \emph{VAS} in Table~\ref{tab:ablationrsb}.
Across all problem sizes and search budgets, we observe a relatively small but consistent improvement ($\sim 1$--$3\%$) from using the remaining search budget $B$ as an explicit input to the policy network.
\vspace{-5pt}
\subsection{Results on the DOTA Dataset}
Next, we repeat our experiments on the DOTA dataset.
We use \textit{large vehicle} and \textit{ship} as our target classes.
In both cases, we also report results with non-overlapping pixel grids of size $200 \times 200$ and $150 \times 150$ ($N=36$ and $N=64$, respectively).
We again use $\mathcal{C} \in \{25, 50, 75\}$.

\begin{table}[h!]
       \scriptsize
        \centering        \caption{\footnotesize{\textbf{ANT} comparisons for the \textit{large vehicle} target class on DOTA.}}
        \vspace{-3pt}
        \begin{tabular}{lcccc}
        \toprule
             \textbf{Method} & $\mathcal{C}=25$ & $\mathcal{C}=50$ & $\mathcal{C}=75$  \\
            \midrule
             \emph{random search} ($N=36$)  & 1.79 & 3.50 & 5.10 \\
             \emph{greedy classification} ($N=36$) & 2.64 & 4.07 & 5.88 \\
             \emph{greedy selection}~\cite{uzkent2020learning} ($N=36$)  & 2.82 & 4.21 & 5.97 \\
             \emph{active learning}~\cite{yoo2019learning} ($N=36$)  & 2.63 & 4.06 & 5.84 \\
             \emph{conventional active search}~\cite{jiang2017efficient} ($N=36$)  & 1.92 & 3.63 & 5.34 \\
             \midrule
             \textit{\textbf{VAS}} ($N=36$) & \textbf{4.63} & \textbf{6.79} & \textbf{8.07} \\
             \midrule
             \midrule
             \emph{random search} ($N=64$)  & 1.48 & 2.96 & 3.91 \\
             \emph{greedy classification} ($N=64$) & 2.59 & 3.77 & 5.48 \\
             \emph{greedy selection}~\cite{uzkent2020learning} ($N=64$)  & 2.72 & 4.10 & 5.77 \\
             \emph{active learning}~\cite{yoo2019learning} ($N=64$)  & 2.57 & 3.74 & 5.47 \\
             \emph{conventional active search}~\cite{jiang2017efficient} ($N=64$)  & 1.64 & 3.15 & 4.23 \\
             \midrule
             \textit{\textbf{VAS}} ($N=64$) & \textbf{5.33} & \textbf{8.47} & \textbf{10.51} \\
             \bottomrule
        \end{tabular}
        \vspace{-4pt}
        \label{tab:dotalv}
\end{table}
\vspace{-4pt}
\begin{table}[h!]
       \scriptsize
        \centering
        \caption{\footnotesize{\textbf{ANT} comparisons for the \textit{ship} target class on the DOTA dataset.}}
        \vspace{-3pt}
        \begin{tabular}{lcccc}
        \toprule
             \textbf{Method} & $\mathcal{C}=25$ & $\mathcal{C}=50$ & $\mathcal{C}=75$  \\
            \midrule
             \emph{random search} ($N=36$)  & 1.73 & 3.07 & 4.26 \\
             \emph{greedy classification} ($N=36$) & 2.04 & 3.65 & 4.92 \\
             \emph{greedy selection}~\cite{uzkent2020learning} ($N=36$)  & 2.33 & 3.84 & 5.01 \\
             \emph{active learning}~\cite{yoo2019learning} ($N=36$)  & 2.01 & 3.64 & 4.91 \\
             \emph{conventional active search}~\cite{jiang2017efficient} ($N=36$)  & 1.86 & 3.25 & 4.40 \\
             \midrule
             \textit{\textbf{VAS}} ($N=36$) & \textbf{3.31} & \textbf{5.34} & \textbf{6.74} \\
             \midrule
             \midrule
             \emph{random search} ($N=64$)  & 1.26 & 2.33 & 3.14 \\
             \emph{greedy classification} ($N=64$) & 1.89 & 3.06 & 3.75 \\
             \emph{greedy selection}~\cite{uzkent2020learning} ($N=64$)  & 2.07 & 3.32 & 4.02 \\
             \emph{active learning}~\cite{yoo2019learning} ($N=64$)  & 1.87 & 3.05 & 3.72 \\
             \emph{conventional active search}~\cite{jiang2017efficient} ($N=64$)  & 1.41 & 2.48 & 3.38 \\
             \midrule
             \textit{\textbf{VAS}} ($N=64$) & \textbf{3.58} & \textbf{6.38} & \textbf{7.83} \\
             \bottomrule
        \end{tabular}
        \label{tab:dotaship}
\end{table}

The results are presented in Tables~\ref{tab:dotalv} and~\ref{tab:dotaship}, and are broadly consistent with our observations on the xView dataset, with \emph{VAS} outperforming all baselines by $\sim 40-80$\%, with the greatest improvement typically coming with a higher search budget $\mathcal{C}$.

\subsection{Visualization of VAS Strategy}

\begin{figure*}
  \centering
  
  \setlength{\tabcolsep}{1pt}
  \renewcommand{\arraystretch}{1}

      
      
      
  
  \includegraphics[width=.98\linewidth]{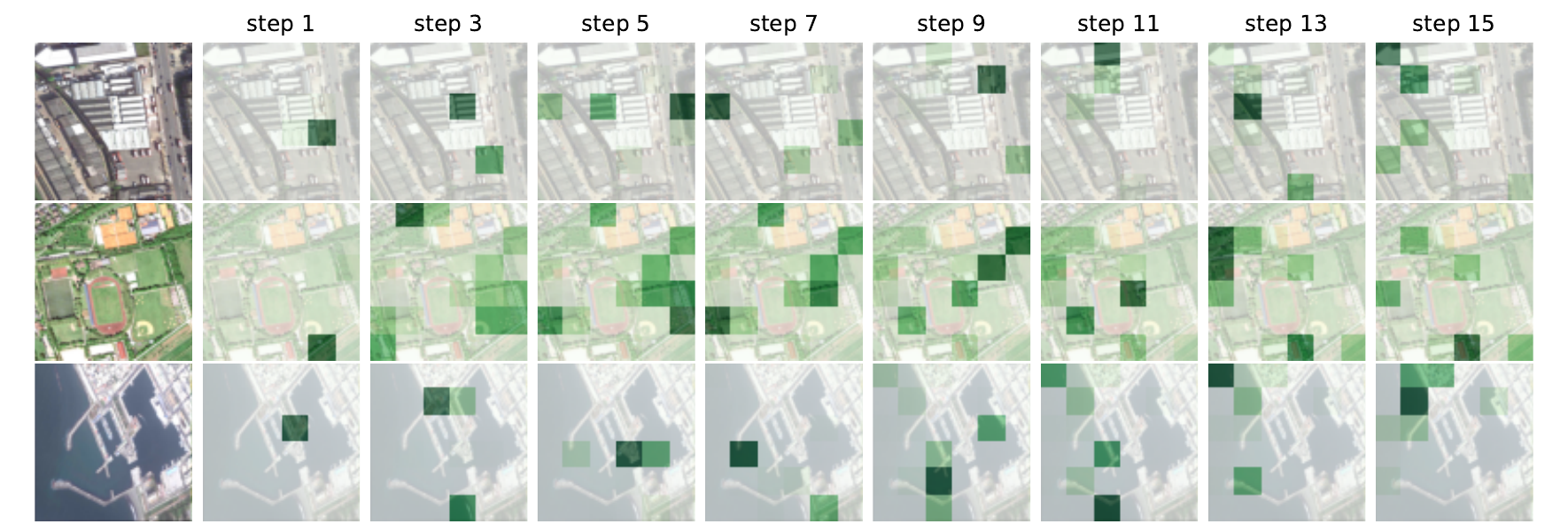}

  \caption{\footnotesize{Query sequences, and corresponding heat maps (darker indicates higher probability), obtained using VAS for different target types.}}

  \label{fig:vis_vas0}
  \vspace{-2pt}
\end{figure*}

In figure~\ref{fig:vis_vas0}, we demonstrate the sequential behavior of a pretrained VAS policy during inference. We have shaded the grid such that darker indicates higher probability. The darkest grid at each step is the target to be revealed. In the first row we have VAS searching for \textit{large vehicles}. In step 1, VAS looks at the roof of a building, which looks very similar to a \textit{large vehicle} in the overhead view. Next in step 3, it searches a grid with a large air conditioner which also looks similar to a \textit{large vehicle}. Having viewed these two confusers, VAS now learns that the rest of the grids with building roofs likely contain no \textit{large vehicles}. It is important to note that this eliminates a large portion of the middle of the image from consideration as it is entirely roof tops. In step 5, it moves to an area which is completely different, a road where it finds a \textit{large vehicle}. VAS now aggressively begins examining grids with roads. In steps 7 through 13 it searches roads discovering \textit{large vehicle}. Finally in step 15 it explores to a parking lot containing a \textit{large vehicle}. In our middle example we have VAS targeting \textit{small cars}. In step 1, VAS targets a road and fails to find a car. In step 3, it searches another road in a different region and finds a car. Having explored regions with prominent major roads it moves to a parking lot in step 5 and finds a car. It now searches a similar parking lot in step 7. Having explored grids with parking lots it goes back to searching minor roads for the duration of its search. VAS does not visit a parking lot in the north east corner, but this parking lot is visually much different from the other two (i.e. it's not rectangular). In our bottom example we have VAS searching for \textit{ships}. In step 1, VAS searches near a harbor. Having found a \textit{ship} it begins exploring similar harbor regions. In step 3 and 5 it searches other parts of the same harbor finding \textit{ships}. In steps 7-9, it searches areas similar to the harbor without \textit{ships}. VAS now learns that \textit{ships} are not likely present in the rest of the dock and explores different regions leaving the rest of the dock unexplored. These three examples demonstrate VAS's tendency for the explore-exploit behavior typical of reinforcement learning algorithms. Additionally, we note that VAS has an ability to eliminate large areas that would otherwise confuse standard greedy approaches.

\subsection{Efficacy of Test-Time Adaptation}

One of the important features of the visual active search problem is that queries actually allow us to observe partial information about target labels \emph{at inference time}.
Here, we evaluate how our approaches to \emph{TTA} that take advantage of this information perform compared to the baseline \emph{VAS without TTA}, as well as state-of-the-art \emph{TTA} baselines discussed in Section~\ref{S:tta} (where \emph{FixMatch} is adapted to also take advantage of observed labels).

\begin{figure}[h!]
    \centering
    \begin{subfigure}[b]{0.23\textwidth}
        \centering
    \includegraphics[width=\textwidth]{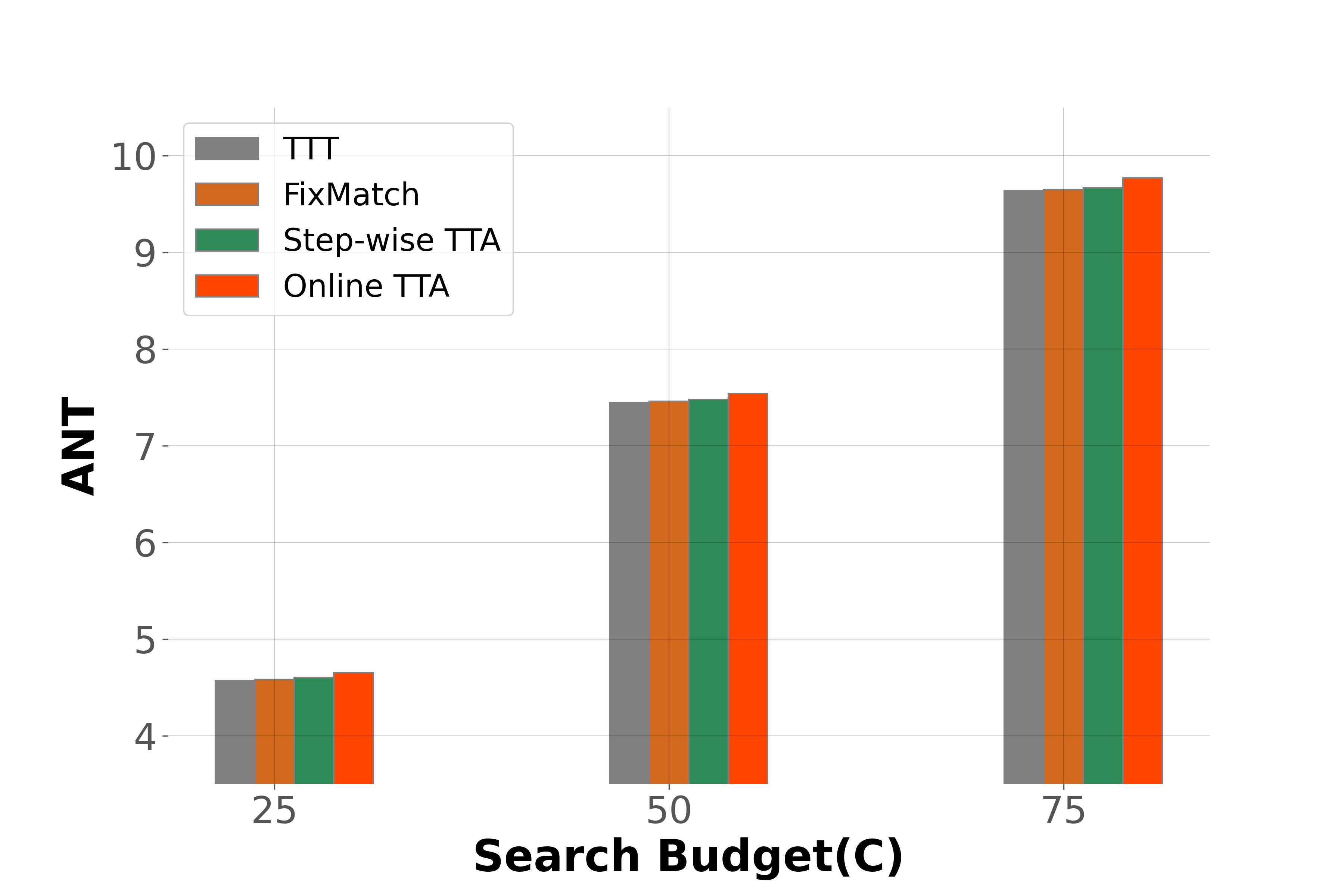}
        \vspace{-3pt}
        \label{fig:esr_sb_48}
    \end{subfigure}%
    \begin{subfigure}[b]{0.23\textwidth}
        \centering
        \includegraphics[width=\textwidth]{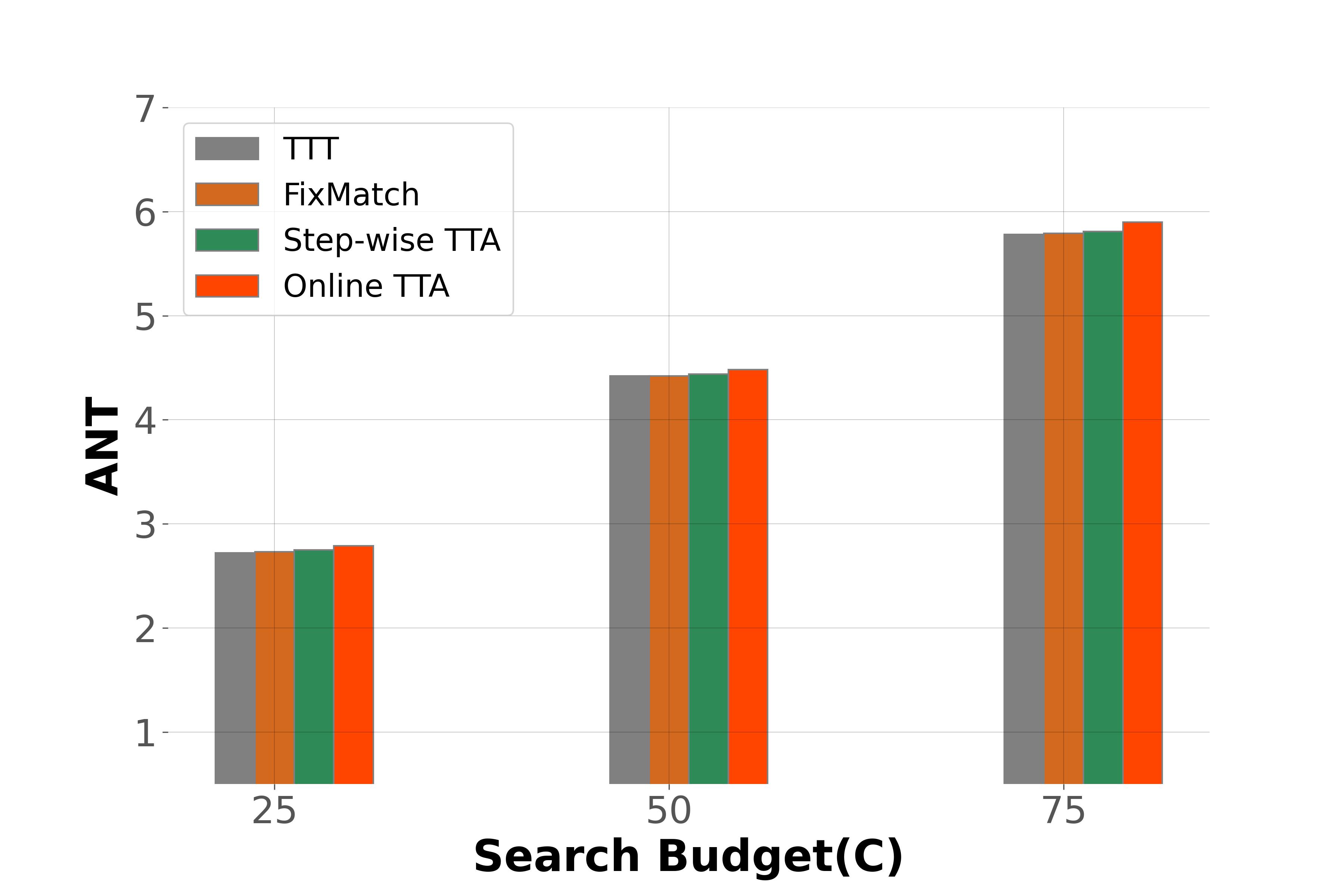}
        \vspace{-3pt}
        \label{fig:esr_tta_99}
    \end{subfigure}
    \begin{subfigure}[b]{0.23\textwidth}
        \centering
    \includegraphics[width=\textwidth]{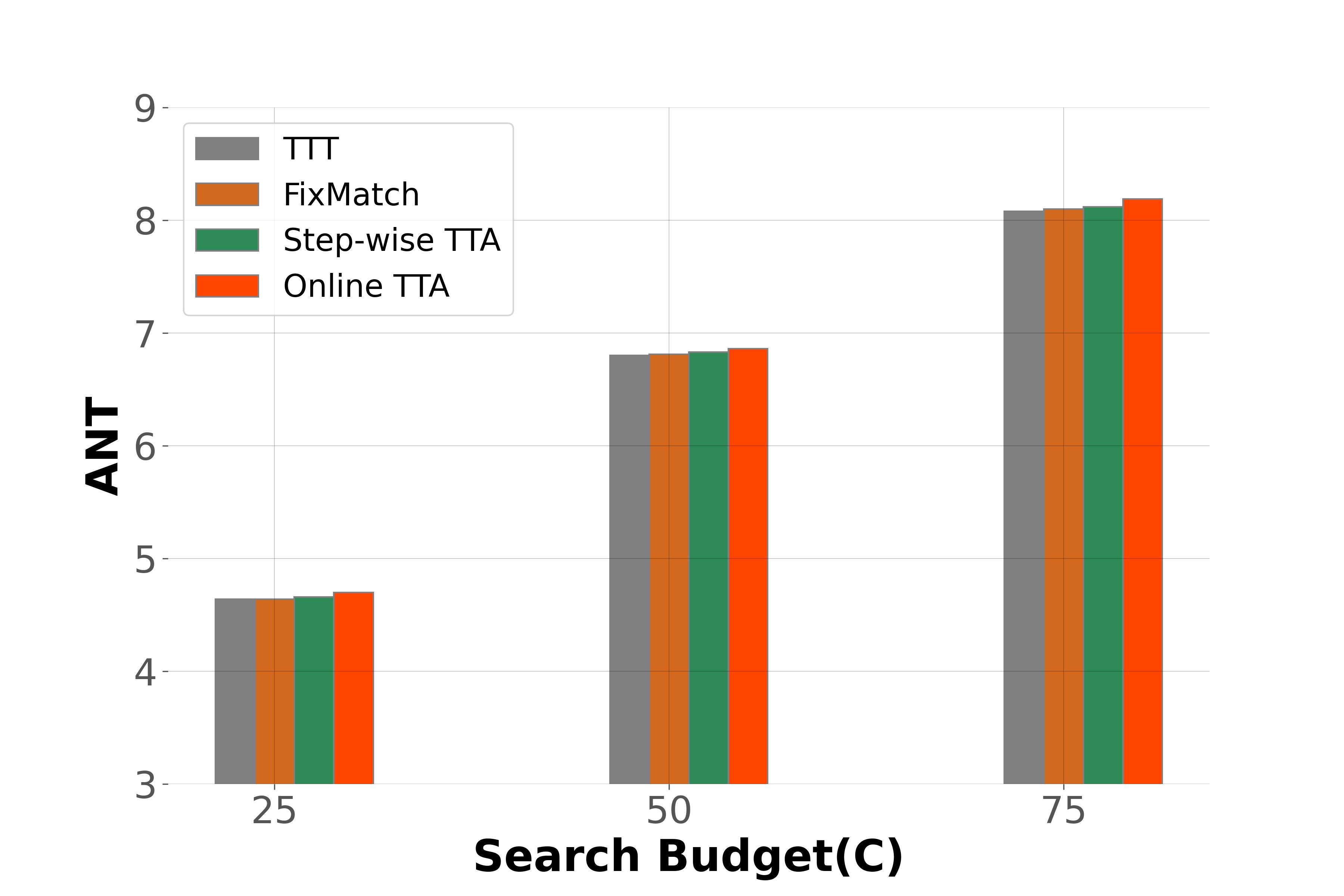}
        \vspace{-3pt}
        \label{fig:esr_sb_36}
    \end{subfigure}%
    \begin{subfigure}[b]{0.23\textwidth}
        \centering
        \includegraphics[width=\textwidth]{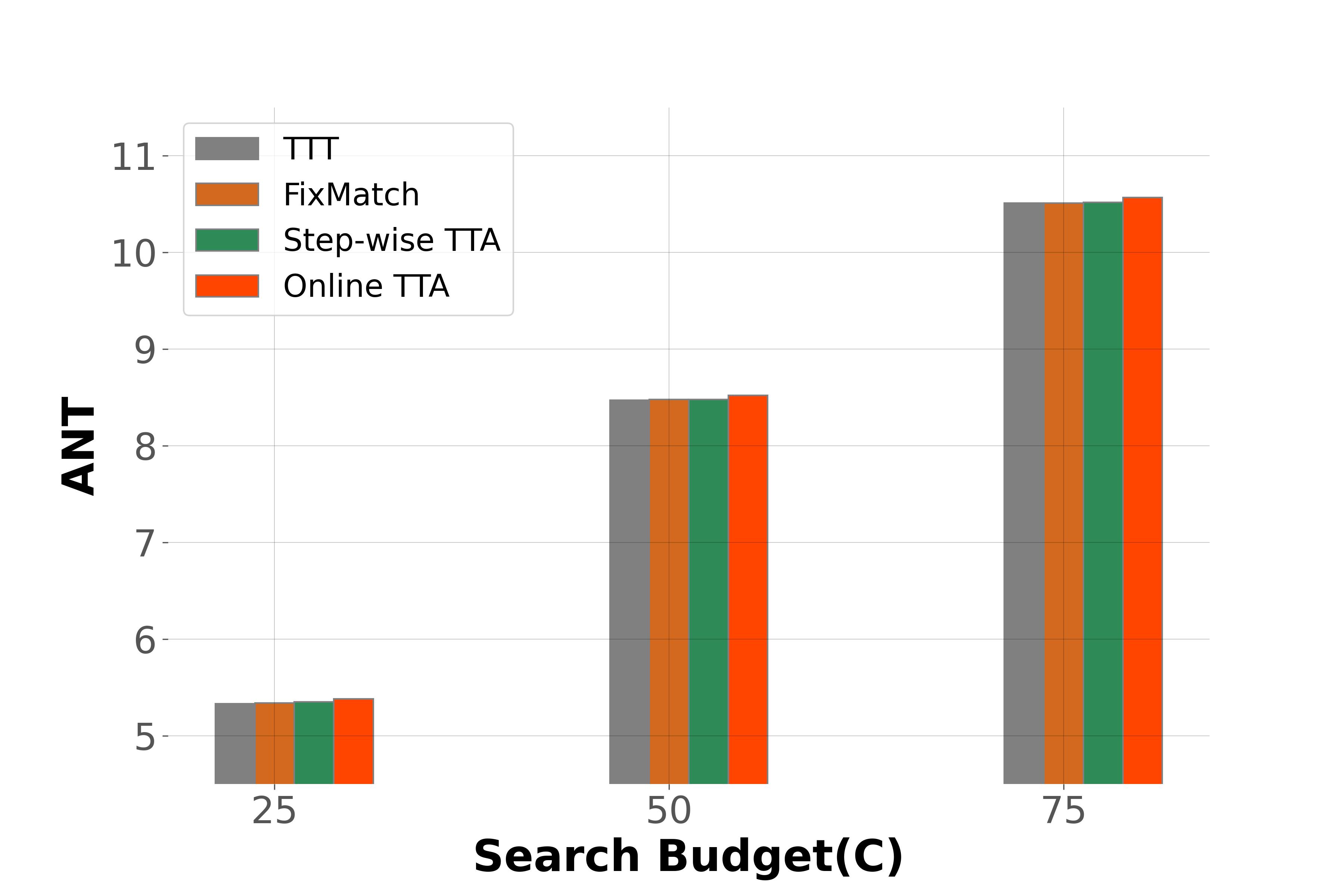}
        \vspace{-3pt}
        \label{fig:esr_tta_64}
    \end{subfigure}
    \caption{\footnotesize{Comparative results of TTA methods on VAS framework. xView (top; \emph{small car} target): (left) $N=48$,  (right) $N=99$. DOTA (bottom; \emph{large vehicle} target): (left) $N=36$, (right) $N=64$.}}
    \vspace{-5pt}
    \label{fig:tta_static}
\end{figure}

Consider first the case where there is no difference between training and test distribution over classes. As before we consider xView and DOTA for analysis. The results are presented in Figure~\ref{fig:tta_static}, and show a consistent pattern.
The \emph{TTT} approach performs the worst, followed by (out adaptation of) \emph{FixMatch}, which is only slightly better than \emph{TTT}.
\emph{Stepwise TTA} outperforms both \emph{TTT} and \emph{FixMatch}, albeit slightly, and \emph{Online TTA} is, somewhat surprisingly much better than all others (this is surprising since it has a lower frequency of model update compared to \emph{Stepwise TTA}).

Finally, we consider a \emph{TTA} setting in which the domain exhibits a non-trivial distributional shift at inference time.
In this case, we would expect the conventional \emph{TTT} and \emph{FixMatch} methods to be more competitive, as they have been specifically designed to account for distribution shift.
We model distribution shift by training the search policy using one target object, and then applying it in the decision context for another target object.
Specifically, for xView, we use \emph{small car} as the target class during training, and \emph{building} as the target class at test time.
Similarly, on the DOTA dataset we use \emph{large vehicle} as the target class at training time, and use \emph{ship} as the target at test time.


\begin{table}[h!]
       \scriptsize
        \centering
        \caption{\footnotesize{Comparative results on xView dataset with \textit{small car} and \textit{Building} as the target class during training and inference respectively.}}
        \begin{tabular}{lcccc}
        \toprule
             \textbf{Method} &  $\mathcal{C}=25$ & $\mathcal{C}=50$ & $\mathcal{C}=75$  \\
            \midrule
             \emph{without TTA} ($N=30$)  & 5.28 & 8.58 & 11.42 \\
             \emph{TTT}~\cite{sun2020test} ($N=30$) & 5.30 & 8.61 & 11.45 \\
             \emph{FixMatch}~\cite{sohn2020fixmatch} ($N=30$)  & 5.31 & 8.62 & 11.47 \\
             \hline
             \textit{\textbf{Stepwise TTA}} ($N=30$)  & 5.33 & 8.64 & 11.50 \\
             \textit{\textbf{Online TTA}} ($N=30$) & \textbf{5.42} & \textbf{8.69} & \textbf{11.58} \\
             \bottomrule
        \end{tabular}
        \vspace{-2pt}
        \label{tab:table_TTA_xview}
\end{table}

\begin{table}[h!]
       \scriptsize
        \centering
        \caption{\footnotesize{Comparative results on DOTA dataset with \textit{large vehicle} and \textit{ship} as the target class during training and inference respectively. }}
        \begin{tabular}{lcccc}
        \toprule
             \textbf{Method} &  $\mathcal{C}=25$ & $\mathcal{C}=50$ & $\mathcal{C}=75$  \\
            \midrule
             \emph{without TTA} ($N=36$)  & 2.69 & 4.38 & 5.84 \\
             \emph{TTT}~\cite{sun2020test} ($N=36$) & 2.70 & 4.39 & 5.84 \\
             \emph{FixMatch}~\cite{sohn2020fixmatch} ($N=36$)  & 2.70 & 4.39 & 5.84  \\
             \hline
             \textit{\textbf{Stepwise TTA}} ($N=36$)  & 2.71 & 4.40  & 5.85 \\
             \textit{\textbf{Online TTA}} ($N=36$) & \textbf{2.73} & \textbf{4.42} & \textbf{5.98} \\
             \bottomrule
        \end{tabular}
        \vspace{-3pt}
        \label{tab:table_TTA_DOTA}
\end{table}
\vspace{-3pt}
The results for the \emph{TTA} setting with distribution shift are presented in Table~\ref{tab:table_TTA_xview} and ~\ref{tab:table_TTA_DOTA} for the xView and the DOTA dataset respectively, where we also add a comparison to the \emph{VAS} without \emph{TTA} of any kind.
We observe that the results here remain consistent, with the proposed \emph{Online TTA} outperforming the other approaches, with \emph{Stepwise TTA} yielding the second-best performance. 
\vspace{-3pt}

%% file: supple.tex
\clearpage
\appendix
\section*{APPENDIX: A Visual Active Search Framework for Geospatial Exploration}
In this appendix, we provide details that could not be included in the main paper owing to space constraints, including: (A) Performance of VAS under uniform query cost; (B) VAS Pseudocode; (C) Policy architecture and training hyperparameter; (D) Search Performance Comparison with Different Feature Extractor Module; (E) More Visual Illustration of VAS and the Most Competitive Greedy Selection baseline Method; (F) Assessment of VAS and other Baseline Methods with a Different Evaluation Metric; (G) Search Performance Comparison with Other Policy Learning Algorithm (PPO); (H) Sensitivity Analysis of \textit{VAS}; (I) Efficacy of TTA on Search Tasks involving Large Number of Grids; (J) Saliency map visualization of \textit{VAS};

\section{Performance of VAS under Uniform Query Cost}
In this section, we report the performance of VAS under uniform query cost. The results are presented in the following Table~\ref{tab:xview-smallcar-sc} for the \textit{small car} target class and in Table~\ref{tab:xview-building-uniform} for the \textit{building} class from the xView dataset. We observe significant improvements in performance of the proposed VAS approach compared to all baselines, ranging from ~$11-25\%$ improvement relative to the most competitive \textit{greedy selection} approach.

\begin{table}[h!]
       \footnotesize
        \centering
        \caption{\textbf{ANT} comparisons for the \textit{small car} target class.}
        \begin{tabular}{lcccc} 
        \toprule
             \textbf{Method} & $\mathcal{C}=12$ & $\mathcal{C}=15$ & $\mathcal{C}=18$  \\
            \midrule
             \emph{random search} ($N=30$)  & 4.57 & 5.66 & 6.85 \\
             \emph{greedy classification} ($N=30$) & 5.31 & 6.24 & 7.25 \\
             \emph{greedy selection}~\cite{uzkent2020learning} ($N=30$)  & 5.47 & 6.45 & 7.46 \\
             \emph{active learning}~\cite{yoo2019learning} ($N=30$)  & 5.28 & 6.21 & 7.22 \\
             \emph{conventional AS}~\cite{jiang2017efficient} ($N=30$)  & 4.86 & 5.97 & 6.92 \\
             \midrule
             \textit{\textbf{VAS}} ($N=30$) & \textbf{6.03} & \textbf{7.24} & \textbf{8.24} \\
             \midrule
             \midrule
             \emph{random search} ($N=48$)  & 3.80 & 4.97 & 5.98 \\
             \emph{greedy classification} ($N=48$) & 4.69 & 5.48 & 6.79 \\
             \emph{greedy selection}~\cite{uzkent2020learning} ($N=48$)  & 4.92 & 5.81 & 6.98 \\
             \emph{active learning}~\cite{yoo2019learning} ($N=48$)  & 4.68 & 5.46 & 6.78 \\
             \emph{conventional AS}~\cite{jiang2017efficient} ($N=48$)  & 3.96 & 5.45 & 6.14 \\
             \midrule
             \textit{\textbf{VAS}} ($N=48$) & \textbf{5.62} & \textbf{6.81} & \textbf{7.86} \\
             \midrule
             \midrule
             \emph{random search} ($N=99$)  & 3.12 & 3.61 & 4.45 \\
             \emph{greedy classification} ($N=99$) & 3.68 & 4.22 & 4.97 \\
             \emph{greedy selection}~\cite{uzkent2020learning} ($N=99$)  & 3.81 & 4.52 & 5.28 \\
             \emph{active learning}~\cite{yoo2019learning} ($N=99$)  & 3.65 & 4.19 & 4.93 \\
             \emph{conventional AS}~\cite{jiang2017efficient} ($N=99$)  & 3.24 & 3.87 & 4.61 \\
             \midrule
             \textit{\textbf{VAS}} ($N=99$) & \textbf{4.61} & \textbf{5.64} & \textbf{6.55} \\
             \bottomrule
        \end{tabular}
        \label{tab:xview-smallcar-sc}
\end{table}

\begin{table}[h!]
       \footnotesize
        \centering
        \caption{\textbf{ANT} comparisons for the \textit{building} target class.}
        \begin{tabular}{lcccc} 
        \toprule
             \textbf{Method} & $\mathcal{C}=12$ & $\mathcal{C}=15$ & $\mathcal{C}=18$  \\
            \midrule
             \emph{random search} ($N=30$)  & 5.54 & 7.18 & 8.58 \\
             \emph{greedy classification} ($N=30$) & 5.88 & 7.72 & 9.21 \\
             \emph{greedy selection}~\cite{uzkent2020learning} ($N=30$)  & 6.39 & 7.95 & 9.52 \\
             \emph{active learning}~\cite{yoo2019learning} ($N=30$)  & 5.86 & 7.68 & 9.16 \\
             \emph{conventional AS}~\cite{jiang2017efficient} ($N=30$)  & 5.76 & 7.37 & 8.87 \\
             \midrule
             \textit{\textbf{VAS}} ($N=30$) & \textbf{7.56} & \textbf{9.02} & \textbf{10.41} \\
             \midrule
             \midrule
             \emph{random search} ($N=48$)  & 4.97 & 6.41 & 7.66 \\
             \emph{greedy classification} ($N=48$) & 5.68 & 6.95 & 8.40 \\
             \emph{greedy selection}~\cite{uzkent2020learning} ($N=48$)  & 5.93 & 7.26 & 8.71 \\
             \emph{active learning}~\cite{yoo2019learning} ($N=48$)  & 5.68 & 6.93 & 8.37 \\
             \emph{conventional AS}~\cite{jiang2017efficient} ($N=48$)  & 5.22 & 6.67 & 7.84 \\
             \midrule
             \textit{\textbf{VAS}} ($N=48$) & \textbf{6.85} & \textbf{8.29} & \textbf{9.65} \\
             \midrule
             \midrule
             \emph{random search} ($N=99$)  & 4.35 & 5.37 & 6.44 \\
             \emph{greedy classification} ($N=99$) & 4.92 & 6.02 & 7.41 \\
             \emph{greedy selection}~\cite{uzkent2020learning} ($N=99$)  & 5.38 & 6.53 & 7.79 \\
             \emph{active learning}~\cite{yoo2019learning} ($N=99$)  & 4.91 & 6.00 & 7.40 \\
             \emph{conventional AS}~\cite{jiang2017efficient} ($N=99$)  & 4.55 & 5.64 & 6.75 \\
             \midrule
             \textit{\textbf{VAS}} ($N=99$) & \textbf{6.75} & \textbf{8.27} & \textbf{9.46} \\
             \bottomrule
        \end{tabular}
        \label{tab:xview-building-uniform}
\end{table}
We also present the results for \textit{large vehicle} and \textit{ship} target class from DOTA dataset in the following Table ~\ref{tab:dotalv-uniform} and ~\ref{tab:dotaship-uniform} respectively. We see the proposed VAS performs noticeably better than all baselines,  ranging from~16--56\% relative to the state-of-the-art greedy selection approach. The experimental outcomes in different settings are qualitatively similar to the settings under Manhattan distance-based query cost.

\begin{table}[h!]
       \footnotesize
        \centering
        \caption{\textbf{ANT} comparisons for the \textit{large vehicle} target class.}
        \begin{tabular}{lcccc}
        \toprule
             \textbf{Method} & $\mathcal{C}=12$ & $\mathcal{C}=15$ & $\mathcal{C}=18$  \\
            \midrule
             \emph{random search} ($N=36$)  & 3.44 & 4.08 & 5.19 \\
             \emph{greedy classification} ($N=36$) & 3.95 & 4.62 & 5.56 \\
             \emph{greedy selection}~\cite{uzkent2020learning} ($N=36$)  & 4.18 & 4.86 & 5.89 \\
             \emph{active learning}~\cite{yoo2019learning} ($N=36$)  & 3.92 & 4.60 & 5.54 \\
             \emph{conventional AS}~\cite{jiang2017efficient} ($N=36$)  & 3.71 & 4.22 & 5.28 \\
             \midrule
             \textit{\textbf{VAS}} ($N=36$) & \textbf{5.14} & \textbf{6.05} & \textbf{7.00} \\
             \midrule
             \midrule
             \emph{random search} ($N=64$)  & 3.40 & 4.03 & 5.14 \\
             \emph{greedy classification} ($N=64$) & 3.87 & 4.59 & 5.55 \\
             \emph{greedy selection}~\cite{uzkent2020learning} ($N=64$)  & 3.99 & 4.77 & 5.67 \\
             \emph{active learning}~\cite{yoo2019learning} ($N=64$)  & 3.85 & 4.54 & 5.51 \\
             \emph{conventional AS}~\cite{jiang2017efficient} ($N=64$)  & 3.61 & 4.12 & 5.26 \\
             \midrule
             \textit{\textbf{VAS}} ($N=64$) & \textbf{6.30} & \textbf{7.65} & \textbf{8.90} \\
             \bottomrule
        \end{tabular}
        \label{tab:dotalv-uniform}
\end{table}
\begin{table}[h!]
       \footnotesize
        \centering
        \caption{\textbf{ANT} comparisons for the \textit{ship} target class.}
        \begin{tabular}{lcccc}
        \toprule
             \textbf{Method} & $\mathcal{C}=12$ & $\mathcal{C}=15$ & $\mathcal{C}=18$  \\
            \midrule
             \emph{random search} ($N=36$)  & 2.69 & 3.38 & 4.46 \\
             \emph{greedy classification} ($N=36$) & 3.21 & 3.99 & 5.11 \\
             \emph{greedy selection}~\cite{uzkent2020learning} ($N=36$)  & 3.44 & 4.23 & 5.32 \\
             \emph{active learning}~\cite{yoo2019learning} ($N=36$)  & 3.18 & 3.95 & 5.07 \\
             \emph{conventional AS}~\cite{jiang2017efficient} ($N=36$)  & 2.97 & 3.56 & 4.77 \\
             \midrule
             \textit{\textbf{VAS}} ($N=36$) & \textbf{4.58} & \textbf{5.34} & \textbf{6.23} \\
             \midrule
             \midrule
             \emph{random search} ($N=64$)  & 2.54 & 3.01 & 4.21 \\
             \emph{greedy classification} ($N=64$) & 3.34 & 3.74 & 4.94 \\
             \emph{greedy selection}~\cite{uzkent2020learning} ($N=64$)  & 3.62 & 3.95 & 5.10 \\
             \emph{active learning}~\cite{yoo2019learning} ($N=64$)  & 3.32 & 3.71 & 4.93 \\
             \emph{conventional AS}~\cite{jiang2017efficient} ($N=64$)  & 2.87 & 3.38 & 4.53 \\
             \midrule
             \textit{\textbf{VAS}} ($N=64$) & \textbf{5.04} & \textbf{6.50} & \textbf{7.38} \\
             \bottomrule
        \end{tabular}
        \label{tab:dotaship-uniform}
\end{table}

\section{VAS Pseudocode}
We have included the pseudocode of our proposed Visual Active Search algorithm in table~\ref{alg:VAS}.

\begin{algorithm}[!]
\small
\caption{The \textsc{VAS} algorithm.}\label{alg:VAS}
\begin{algorithmic}[1]  
\Require  A search task instance $(x_i, y_i)$; budget constraint $\mathcal{C}$; search policy $\psi(x_i, o, B)$ with parameters $\theta$; 
\State \textbf{Initialize} $o^{0} = [ 0...0]$; $B^{0} = \mathcal{C}$; step $t = 0$
\While{$B^{t} > 0 $}
    
    \State $\tilde{y} = \psi(x_i, o^{t}, B^{t})$
    \State \emph{j} $\xleftarrow{} \mathit{Sample}_{j \in \{\mathit{Unexplored\>Grids}\}} [\tilde{y}]$
    \State Query grid cell with index $j$ and observe true label $y^{(j)}$.
    \State Obtain reward $R^t = y^{(j)}$.
    \State Update $o^{t}$ to $o^{t+1}$ with $o^{(j)} = 2y^{(j)} -1$.
    \State Update $B^{t}$ to $B^{t+1}$ with $B^{t+1} = B^{t} - c(k,j)$ (assuming we query $k$'th grid at $(t-1)$).
    \State Collect transition tuple ($\tau$) at step t, i.e., $\tau^{t}$ = $($ state = $(x_i, o^{t},B^{t})$, action = $j$, reward = $R^{t}$, next state = $(x_i, o^{t+1},B^{t+1})$ $)$.
    \State \emph{t} $\xleftarrow{} t+1$
\EndWhile
\State Update the search policy parameters, i.e., $\theta$  using \textit{REINFORCE} objective as in ~\ref{eq:REINFORCE} based on the collected transition tuples ($\tau^{t}$) throughout the episode.
\State \textbf{Return} updated search policy parameters, i.e., $\theta$.
\end{algorithmic}
\label{A:VAS}
\end{algorithm}

\section{Policy architecture, training hyperparameter, and the details of TTA}

In table~\ref{tab:table_ARCH}, we detail the \textit{VAS} policy architecture with number of target grids as $N$. We use a learning rate of $10^{-4}$, batch size of 16, number of training epochs 200, and the Adam optimizer to train the policy network in all results.
We add a self-supervised head $r$ to the VAS policy architecture for TTT. The architecture of self-supervised head is detailed in table~\ref{tab:table_ARCH_recon}. We applied a series of 4 up-convolution layers with intermediate ReLU activations followed by a tanh activation layer on the semantic features extracted using ResNet34. 
For FixMatch, our VAS architecture remains unchanged, and we apply only spatially invariant augmentations (e.g auto contrast, brightness, color, and contrast) and ignore all translation augmentations (translate X, translate Y, ShearX  etc.) to obtain the augmented version of the input image. We update the model parameters after every query step using a cross-entropy loss between a pseudo-target and a predicted vector as described below. We define the pseudo-target vector as follows. Whenever a query $j$ is successful ($y_j = 1$), we construct a label vector as the one-hot vector with a 1 in the $j$th grid cell. However if $y_j=0$, we associate each queried grid cell with a 0, and assign a uniform probability distribution over all unqueried grids. Prediction vector is the “logit” representation obtained from the VAS policy. 
We used the Adam optimizer with a learning rate of $10^{-4}$ for both TTT and FixMatch.

\begin{table}[h!]
        \footnotesize
        \centering
        \caption{VAS Policy Architecture}
        \begin{tabular}{lcccc}
        \toprule
             \textbf{Layers} &  \textbf{Configuration} & \textbf{o/p Feature Map size}   \\
            \midrule
             Input  & RGB Image & 3 $\times$ 2500 $\times$ 3000   \\
             \midrule            
             Feat. Extraction & ResNet-34 & 512 $\times$ 14 $\times$ 14  \\
             \midrule            
             Conv1 & c:N\>\>\>k:$1 \times 1$ & N $\times$ 14 $\times$ 14  \\
             \midrule
             Tile1  & Grid State ($o$) & $N \times 14 \times 14$  \\
             \midrule
             Tile2  & Query Left ($B$)  & $1 \times 14 \times 14$  \\
             \midrule
             Channel Concat  & Conv1,Tile1,Tile2  & $(2N+1) \times 14 \times 14$  \\
             \midrule
             Conv2 & c:3\>\>\>k:$1 \times 1$ & 3 $\times$ 14 $\times$ 14  \\
             \midrule
             Flattened & Conv2 & 588  \\
             \midrule
             FC1+ReLU & ($588 -> 2N$) & 2N  \\
             \midrule
             FC2 & ($2N -> N$) & N  \\
             \bottomrule
        \end{tabular}
        \label{tab:table_ARCH}
\end{table}

\begin{table}[h!]
        \scriptsize
        \centering
        \caption{Self-supervised head Architecture}
        \begin{tabular}{lcc}
        \toprule
             \textbf{Layers} &  \textbf{Configuration}   \\
            \midrule
             Input: Latent Feature  & $36 \times 14 \times 14$   \\
             \midrule            
             1st Up-conv layer & in-channel:36;out-channel:36;k:$ 3\times 3$;stride:2;padd:0    \\
             \midrule            
             Activation Layer & ReLU  \\
             \midrule
             2nd Up-conv layer  & in-channel:36;out-channel:24;k:$ 3\times 3$;stride:2;padd:1   \\
             \midrule
             Activation Layer & ReLU   \\
             \midrule
             3rd Up-conv layer  & in-channel:24;out-channel:12;k:$ 2\times 2$;stride:4;padd:1    \\
             \midrule
             Activation Layer & ReLU  \\
             \midrule
             4th Up-conv layer  & in-channel:12;\>out-channel:3;\>k:$ 2\times 2$;\>stride:2;\>padd:0 \\
             \midrule
             Normalization layer & tanh  \\
             \bottomrule
        \end{tabular}
        \label{tab:table_ARCH_recon}
\end{table}

\section{Search Performance Comparison with Different Feature Extractor Module}
In this section, we compare the performance of VAS with different feature extraction module. We use state-of-the-art feature extraction modules, such as ViT~\cite{dosovitskiy2020image} and DINO~\cite{caron2021emerging} for comparison. The Vision Transformer (ViT)~\cite{dosovitskiy2020image} is a transformer encoder model (BERT-like) pretrained on a large collection of images in a self-supervised fashion, namely ImageNet-21k (a collection of 14 million images), at a resolution of $224\times224$ pixels, with patch resolution of $16\times16$. Note that, we use off the shelf pretrained ViT model provided by huggingface (google/vit-base-patch16-224-in21k). We call the resulting policy \textit{VAS-ViT}. Similar to ViT, DINO~\cite{caron2021emerging} is also based on transformer encoder model. Images are presented to the DINO model as a sequence of fixed-size patches (resolution 8x8), which are linearly embedded. For our experiment, we use DINO pretrained on ImageNet-1k, at a resolution of 224x224 pixels. For our experiments, we use pretrained DINO model provided by huggingface (facebook/dino-vits8). We call the resulting policy as VAS-DINO. In table ~\ref{tab:xview-smallcar_FE}, ~\ref{tab:dotalv_FE} we report the performance of VAS-ViT and VAS-DINO and compare them with VAS.

\begin{table}[h!]
       \small
        \centering
        \caption{\small{\textbf{ANT} comparisons with different feature extraction module for the \textit{small car} target class on xView.}}
        \begin{tabular}{lcccc}
        \toprule
             \textbf{Method} & $\mathcal{C}=25$ & $\mathcal{C}=50$ & $\mathcal{C}=75$  \\
            \midrule
             \emph{VAS-DINO} ($N=30$)  & 4.56 & 7.41 & 9.83 \\
             \emph{VAS-ViT} ($N=30$)  & \textbf{4.64} & 7.47 & 9.86 \\
             \midrule
             \textit{\textbf{VAS}} ($N=30$) & 4.61 & \textbf{7.49} & \textbf{9.88} \\
             \midrule
             \midrule
             \emph{VAS-DINO} ($N=48$)  & 4.52 & 7.41 & 9.59 \\
             \emph{VAS-ViT} ($N=48$)  & 4.56 & 7.44 & \textbf{9.68} \\
             \midrule
             \textit{\textbf{VAS}} ($N=48$) & \textbf{4.56} & \textbf{7.45} & 9.63 \\
             \midrule
             \bottomrule
        \end{tabular}
        \label{tab:xview-smallcar_FE}
\end{table}

\begin{table}[h!]
       \small
        \centering
        \caption{\small{\textbf{ANT} comparisons with different feature extraction module for the \textit{large vehicle} target class on DOTA.}}
        \vspace{-3pt}
        \begin{tabular}{lcccc}
        \toprule
             \textbf{Method} & $\mathcal{C}=25$ & $\mathcal{C}=50$ & $\mathcal{C}=75$  \\
            \midrule
             \emph{VAS-DINO} ($N=36$)  & 4.56 & 6.75 & 8.03 \\
             \emph{VAS-ViT} ($N=36$) & 4.60 & \textbf{6.82} & \textbf{8.09} \\
             \midrule
             \textit{\textbf{VAS}} ($N=36$) & \textbf{4.63} & 6.79 & 8.07 \\
             \midrule
             \midrule
             \emph{VAS-DINO} ($N=64$)  & 5.27 & 8.44 & 10.45 \\
             \emph{VAS-ViT} ($N=64$) & 5.31 & \textbf{8.51} & 10.48 \\
             \midrule
             \textit{\textbf{VAS}} ($N=64$) & \textbf{5.33} & 8.47 & \textbf{10.51} \\
             \bottomrule
        \end{tabular}
        \label{tab:dotalv_FE}
\end{table}

\section{More Visual Illustration of VAS and the Most Competitive Greedy Selection baseline Method}
\begin{figure}[h!]
    \centering
    \begin{subfigure}[b]{0.23\textwidth}
        \centering
    \includegraphics[width=.95\linewidth]{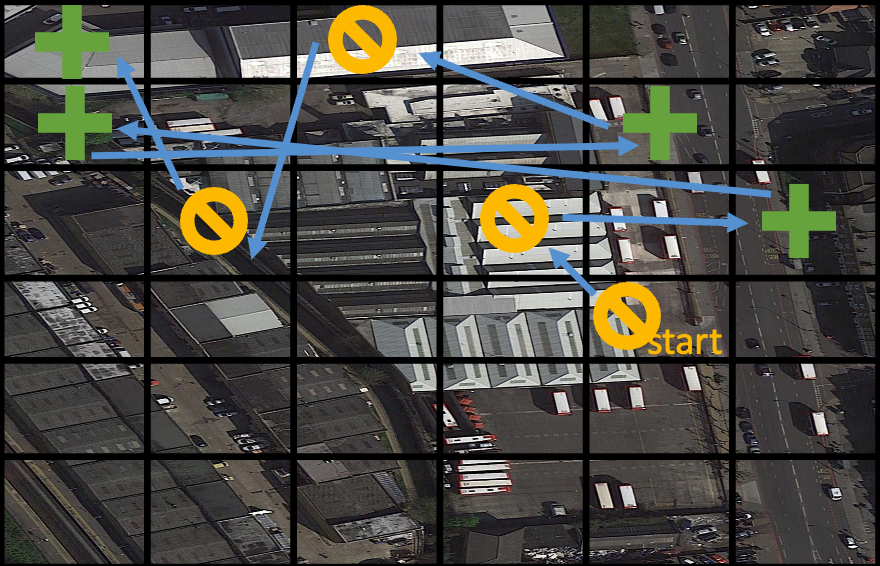}
        \label{fig:vis_vas}
    \end{subfigure}%
    \begin{subfigure}[b]{0.23\textwidth}
        \centering
    \includegraphics[width=.95\linewidth]{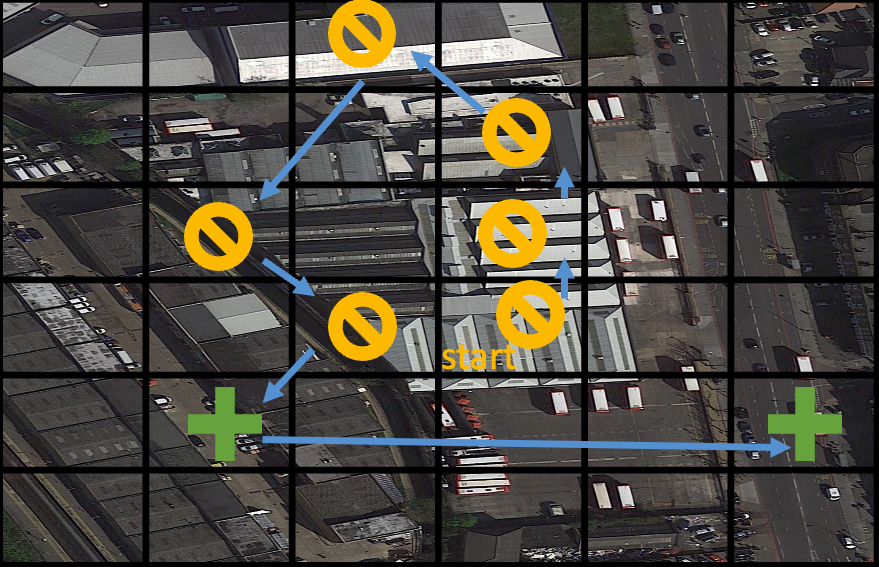}
        \label{fig:vis_greedy}
    \end{subfigure}
    \caption{\footnotesize{Comparison of policies learned using \textit{VAS} (left) and the \emph{greedy selection} baseline method (right).} 
    }
    \vspace{-2pt}
    \label{fig:vis_vas_r1}
\end{figure}

\begin{figure}[h!]
    \centering
    \begin{subfigure}[b]{0.23\textwidth}
        \centering
    \includegraphics[width=.95\linewidth]{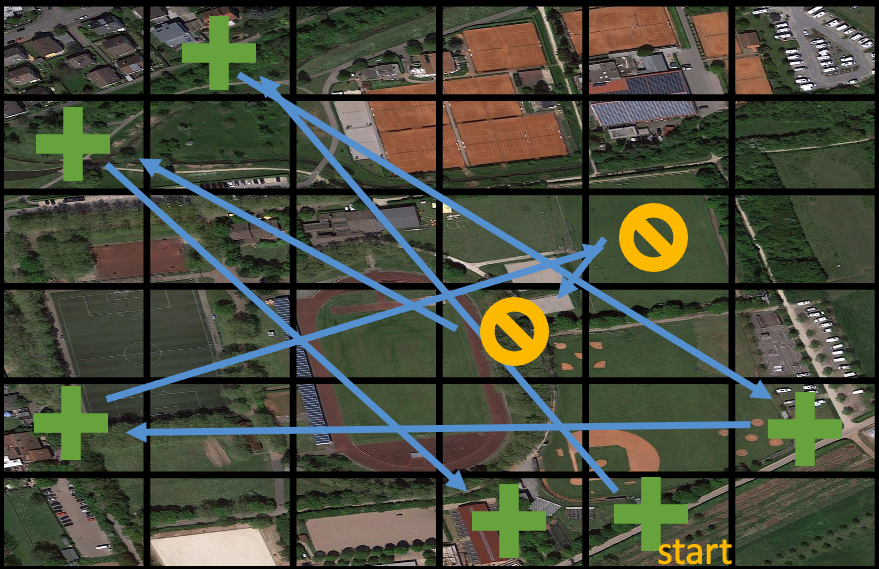}
        \label{fig:vis_vas}
    \end{subfigure}%
    \begin{subfigure}[b]{0.23\textwidth}
        \centering
    \includegraphics[width=.95\linewidth]{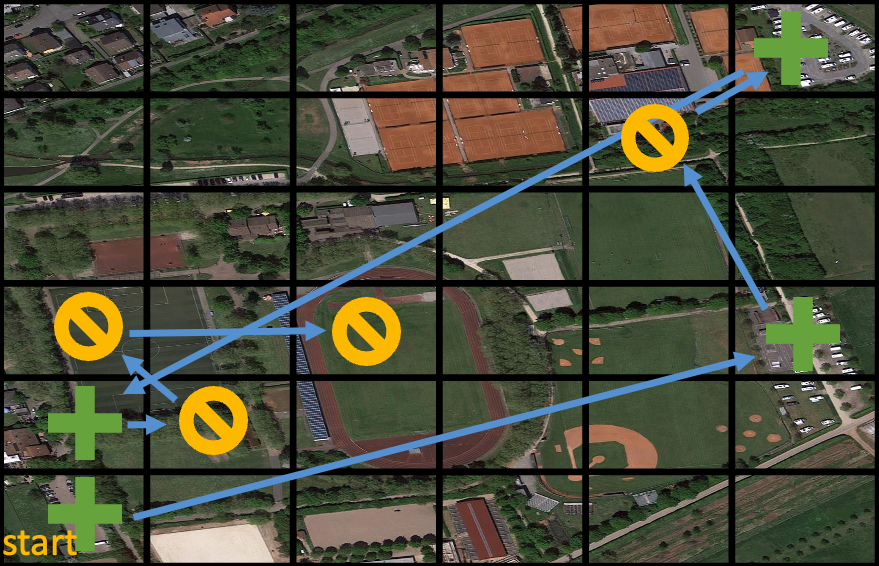}
        \label{fig:vis_greedy}
    \end{subfigure}
    \caption{\footnotesize{Comparison of policies learned using \textit{VAS} (left) and the \emph{greedy selection} baseline method (right).} 
    }
    \vspace{-2pt}
    \label{fig:vis_vas_r2}
\end{figure}

\begin{figure}[h!]
    \centering
    \begin{subfigure}[b]{0.23\textwidth}
        \centering
    \includegraphics[width=.95\linewidth]{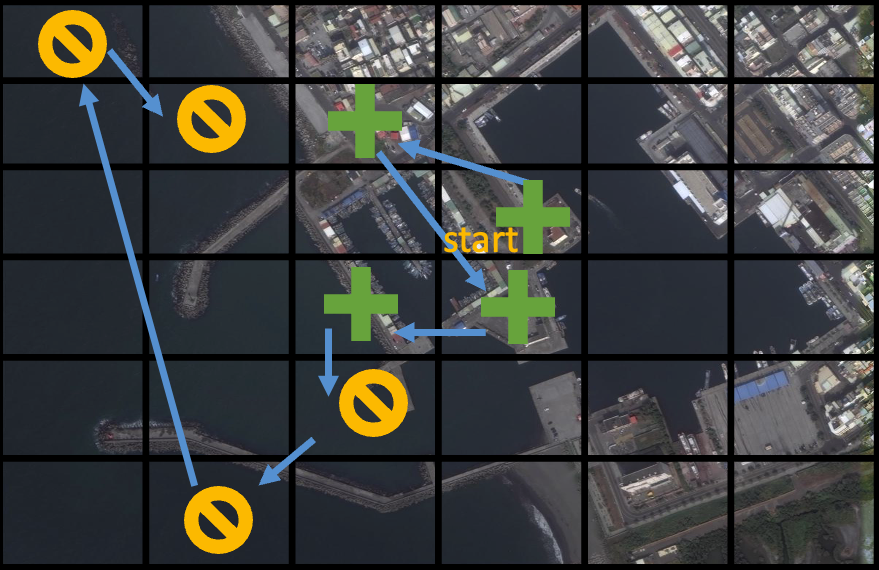}
        \label{fig:vis_vas}
    \end{subfigure}%
    \begin{subfigure}[b]{0.23\textwidth}
        \centering
    \includegraphics[width=.95\linewidth]{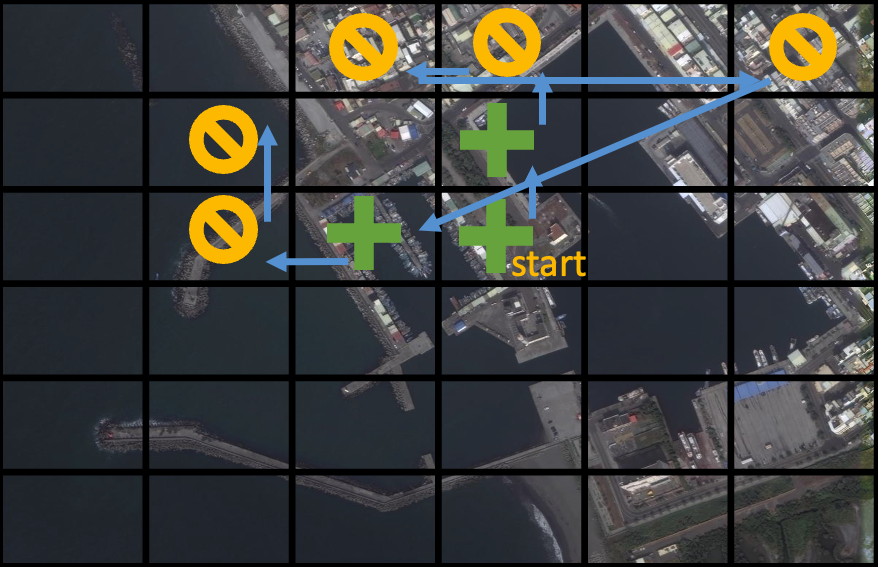}
        \label{fig:vis_greedy}
    \end{subfigure}
    \caption{\footnotesize{Comparison of policies learned using \textit{VAS} (left) and the \emph{greedy selection} baseline method (right).} 
    }
    \vspace{-2pt}
    \label{fig:vis_vas_r3}
\end{figure}

In this section, we provide additional visualization of comparative exploration behaviour of VAS and and the most competitive greedy selection baseline approach. In figure ~\ref{fig:vis_vas_r1}, we compare the search strategy with \emph{large vehicle} as a target class. In figure ~\ref{fig:vis_vas_r2}, we compare the behaviour with \emph{small car} as a target class. In figure ~\ref{fig:vis_vas_r3}, we analyze the exploration behaviour with \emph{ship} as a target class.

These additional visualizations again justify the efficacy of VAS over the strongest baseline method.
\section{Assessment of VAS and other Competitive Baseline Methods with a Different Evaluation Metric}
We additionally compare the search performance of VAS with all the baseline methods using a metric, which we call \textit{Effective Success Rate (ESR)}. A na\"ive way to evaluate the proposed approaches is to simply use \emph{success rate}, which is is the fraction of total search steps $K$ that identify a target object.
However, if $K$ exceeds the total number of target objects in $x$, normalizing by $K$ is unreasonable, as even a perfect search strategy would appear to work poorly.
Consequently, we propose \emph{effective success rate (ESR)} as the efficacy metric, defined as follows:
\begin{align}
   \text{\textbf{ESR}} =
   \frac{\# \>Targets\>Discovered}{\min\{\# \>Targets\>, K\}}
   \label{eq:metric_esr}
\end{align}
Thus, we divide by the number of targets one can possibly discover given a search budget $K$, rather than simply the search budget.

\subsection{Results on the xView Dataset with ESR as Evaluation Metric}
We initiate our analysis by assessing the proposed methodologies using the xView dataset, for varying search budgets $K \in \{12, 15, 18\}$ and number of grid cells $N \in \{30, 48, 99\}$.
We also consider two target classes for our search: \emph{small car} and \emph{building}.
As the dataset contains variable size images, take random crops of $2500 \times 3000$ for $N=30$, $2400 \times 3200$ pixels for $N=48$, and $2700 \times 3300$ for $N=99$, thereby guarantees uniform grid cell dimensions across the board.

\begin{table}[h!]
       \footnotesize
        \centering
        \caption{\textbf{ESR} comparisons for the \textit{small car} target class on the xView dataset.}
        \begin{tabular}{lcccc}
        \toprule
             \textbf{Method} & $K=12$ & $K=15$ & $K=18$  \\
            \midrule
             \emph{random search} ($N=30$)  & 0.598 & 0.632 & 0.704 \\
             \emph{greedy classification} ($N=30$) & 0.619 & 0.675 & 0.718 \\
             \emph{greedy selection}~\cite{uzkent2020learning} ($N=30$)  & 0.627 & 0.684 & 0.729 \\
            \midrule
              \textit{\textbf{VAS}} ($N=30$) & \textbf{0.766} & \textbf{0.826} & \textbf{0.861} \\
            \midrule
             \midrule
             \emph{random search} ($N=48$)  & 0.489 & 0.517 & 0.558 \\
             \emph{greedy classification} ($N=48$) & 0.512 & 0.551 & 0.589 \\
             \emph{greedy selection}~\cite{uzkent2020learning} ($N=48$)  & 0.524 & 0.568 & 0.596 \\
            \midrule
             \textit{\textbf{VAS}} ($N=48$) & \textbf{0.694}  & \textbf{0.722} & \textbf{0.741} \\
            \midrule
            \midrule
             \emph{random search} ($N=99$)  & 0.336 & 0.369 & 0.378 \\
             \emph{greedy classification} ($N=99$) & 0.365 & 0.384 & 0.405  \\
             \emph{greedy selection}~\cite{uzkent2020learning} ($N=99$)   & 0.376 & 0.395 & 0.418 \\
             \midrule
             \textit{\textbf{VAS}} ($N=99$)  & \textbf{0.564}  & \textbf{0.587} & \textbf{0.602} \\
             \bottomrule
        \end{tabular}
        \label{tab:xview-smallcar_ESR}
\end{table}

\begin{table}[h!]
       \footnotesize
        \centering
        \caption{\textbf{ESR} comparisons for the \textit{building} target class on the xView dataset. }
        \begin{tabular}{lcccc}
        \toprule
             \textbf{Method} &  $K=12$ & $K=15$ & $K=18$  \\
            \midrule
             \emph{random search} ($N=30$)  & 0.663 & 0.681 & 0.697 \\
             \emph{greedy classification} ($N=30$) & 0.701 & 0.734 & 0.767 \\
             \emph{greedy selection}~\cite{uzkent2020learning} ($N=30$)  & 0.708 & 0.740 & 0.786 \\
             \midrule
             \textit{\textbf{VAS}} ($N=30$) & \textbf{0.854} & \textbf{0.886} & \textbf{0.912} \\
             \midrule
             \midrule
             \emph{random search} ($N=48$)  & 0.526 & 0.547 & 0.556 \\
             \emph{greedy classification} ($N=48$) & 0.548 & 0.569 & 0.585 \\
             \emph{greedy selection}~\cite{uzkent2020learning} ($N=48$)  & 0.552 & 0.574 & 0.604 \\
             \midrule
             \textit{\textbf{VAS}} ($N=48$) & \textbf{0.677} & \textbf{0.716} & \textbf{0.738} \\
             \midrule
             \midrule
             \emph{random search} ($N=99$)  & 0.443 & 0.462 & 0.483 \\
             \emph{greedy classification} ($N=99$) & 0.460 & 0.482 & 0.504 \\
             \emph{greedy selection}~\cite{uzkent2020learning}  & 0.469 & 0.488 & 0.514 \\
             \midrule
             \textit{\textbf{VAS}} ($N=99$)  & \textbf{0.654}  & \textbf{0.676} & \textbf{0.690} \\
             \bottomrule
        \end{tabular}
        \label{tab:xview-building_ESR}
\end{table}

The results are presented in Table~\ref{tab:xview-smallcar_ESR} for the \emph{small car} class and in Table~\ref{tab:xview-building_ESR} for the \emph{building} class.
We see significant improvements in performance of the proposed \emph{VAS} approach compared to all baselines, ranging from $\sim$15--50\% improvement relative to the most competitive state-of-the-art method, \emph{greedy selection}.
\subsection{Results on the DOTA Dataset with ESR as Evaluation Metric}

We also conduct our experiments on the DOTA dataset.
We use \textit{large vehicle} and \textit{ship} as our target classes.
In both cases, we also report results with non-overlapping pixel grids of size $200 \times 200$ and $150 \times 150$ ($N=36$ and $N=64$, respectively).
We again use $K\in \{12, 15, 18\}$.

\begin{table}[h!]
       \footnotesize
        \centering
        \caption{\textbf{ESR} comparisons for the \textit{large vehicle} target class on the DOTA dataset.}
        \begin{tabular}{lcccc}
        \toprule
             \textbf{Method} & $K=12$ & $K=15$ & $K=18$  \\
            \midrule
             \emph{random search} ($N=36$)  & 0.460 & 0.498 & 0.533 \\
             \emph{greedy classification} ($N=36$) & 0.602 & 0.624 & 0.641 \\
             \emph{greedy selection}~\cite{uzkent2020learning} ($N=36$)  & 0.618 & 0.637 & 0.647 \\
             \midrule
             \textit{\textbf{VAS}} ($N=36$) & \textbf{0.736} & \textbf{0.744} & \textbf{0.767} \\
             \midrule
             \midrule
             \emph{random search} ($N=64$)  & 0.389 & 0.405 & 0.442 \\
             \emph{greedy classification} ($N=64$) & 0.606 & 0.612 & 0.618 \\
             \emph{greedy selection}~\cite{uzkent2020learning} ($N=64$)  & 0.612 & 0.618 & 0.626 \\
             \midrule
             \textit{\textbf{VAS}} ($N=64$) & \textbf{0.724} & \textbf{0.738} & \textbf{0.749} \\
             \bottomrule
        \end{tabular}
        \label{tab:dotalv_ESR}
\end{table}

\begin{table}[h!]
       \footnotesize
        \centering
        \caption{\textbf{ESR} comparisons for the  \textit{ship} target class on the DOTA dataset.}
        \begin{tabular}{lcccc}
        \toprule
             \textbf{Method} & $K=12$ & $K=15$ & $K=18$  \\
            \midrule
             \emph{random search} ($N=36$)  & 0.491 & 0.564 & 0.590 \\
             \emph{greedy classification}($N=36$) & 0.602 & 0.629 & 0.657 \\
            \emph{greedy selection}~\cite{uzkent2020learning} ($N=36$)  & 0.609 & 0.638 & 0.665 \\
             \midrule
             \textit{\textbf{VAS}} ($N=36$) & \textbf{0.757} & \textbf{0.764} & \textbf{0.776} \\
             \midrule
             \midrule
             \emph{random search} ($N=64$)  & 0.334 & 0.379 & 0.417 \\
             \emph{greedy classification} ($N=64$) & 0.524 & 0.541 & 0.559 \\
             \emph{greedy selection}~\cite{uzkent2020learning} ($N=64$)  & 0.531 & 0.552 & 0.576 \\
             \midrule
             \textit{\textbf{VAS}} ($N=64$) & \textbf{0.700} & \textbf{0.712} & \textbf{0.733} \\
             \bottomrule
        \end{tabular}
        \label{tab:dotaship_ESR}
\end{table}
The results are presented in Tables~\ref{tab:dotalv_ESR} and~\ref{tab:dotaship_ESR}, and are broadly consistent with our observations on the xView dataset, with \emph{VAS} outperforming all baselines by $\sim$16--25\%, with the greatest improvement typically coming on more difficult tasks (small $K$ compared to $N$).

\section{Search Performance Comparison with Other Policy Learning Algorithm (PPO)}
We conduct experiments with other policy learning algorithm, such as PPO. With PPO~\cite{schulman2017proximal}, the idea is to constrain our policy update with a new objective function called the clipped surrogate objective function that will constrain the policy change in a small range $[1 - \epsilon, 1 + \epsilon]$. Here, $\epsilon$ is a hyperparameter that helps us to define this clip range. In all our experiment with PPO, we use clip range $\epsilon = 0.2$ as provided in the main paper~\cite{schulman2017proximal}. We keep all other hyperparameters including policy architecture fixed. We call the resulting policy \emph{VAS-PPO}. In table ~\ref{tab:xview-smallcar_PPO},~\ref{tab:dotalv_PPO} we present the result of VAS-PPO and compare the performance with VAS. our experimental finding suggests that PPO doesn't yield any extra benefits in spite of having added complexity overhead due to the clipped surrogate objective. 

\begin{table}[h!]
       \small
        \centering
        \caption{\small{\textbf{ANT} comparisons with different policy learning algorithm for the \textit{small car} target class on xView.}}
        \begin{tabular}{lcccc}
        \toprule
             \textbf{Method} & $\mathcal{C}=25$ & $\mathcal{C}=50$ & $\mathcal{C}=75$  \\
            \midrule
             \emph{VAS-PPO} ($N=30$)  & 4.15 & 6.82 & 9.16 \\
             \midrule
             \textit{\textbf{VAS}} ($N=30$) & \textbf{4.61} & \textbf{7.49} & \textbf{9.88} \\
             \midrule
             \midrule
             \emph{VAS-PPO} ($N=48$)  & 4.03 &6.87 & 9.02 \\
             \midrule
             \textit{\textbf{VAS}} ($N=48$) & \textbf{4.56} & \textbf{7.45} & \textbf{9.63} \\
             \midrule
             \bottomrule
        \end{tabular}
        \label{tab:xview-smallcar_PPO}
\end{table}

\begin{table}[h!]
       \small
        \centering
        \caption{\small{\textbf{ANT} comparisons with different policy learning algorithm for the \textit{large vehicle} target class on DOTA.}}
        \vspace{-3pt}
        \begin{tabular}{lcccc}
        \toprule
             \textbf{Method} & $\mathcal{C}=25$ & $\mathcal{C}=50$ & $\mathcal{C}=75$  \\
            \midrule
             \emph{VAS-PPO} ($N=36$) & 4.01 & 6.24 & 7.56 \\
             \midrule
             \textit{\textbf{VAS}} ($N=36$) & \textbf{4.63} & \textbf{6.79} & \textbf{8.07} \\
             \midrule
             \midrule
             \emph{VAS-PPO} ($N=64$) & 4.89 & 7.93 & 10.12 \\
             \midrule
             \textit{\textbf{VAS}} ($N=64$) & \textbf{5.33} & \textbf{8.47} & \textbf{10.51} \\
             \bottomrule
        \end{tabular}
        \label{tab:dotalv_PPO}
\end{table}


\section{Sensitivity Analysis of \textit{VAS}}
\begin{figure*}
\centering
\begin{subfigure}[b]{0.50\textwidth}
   \includegraphics[width=1\linewidth]{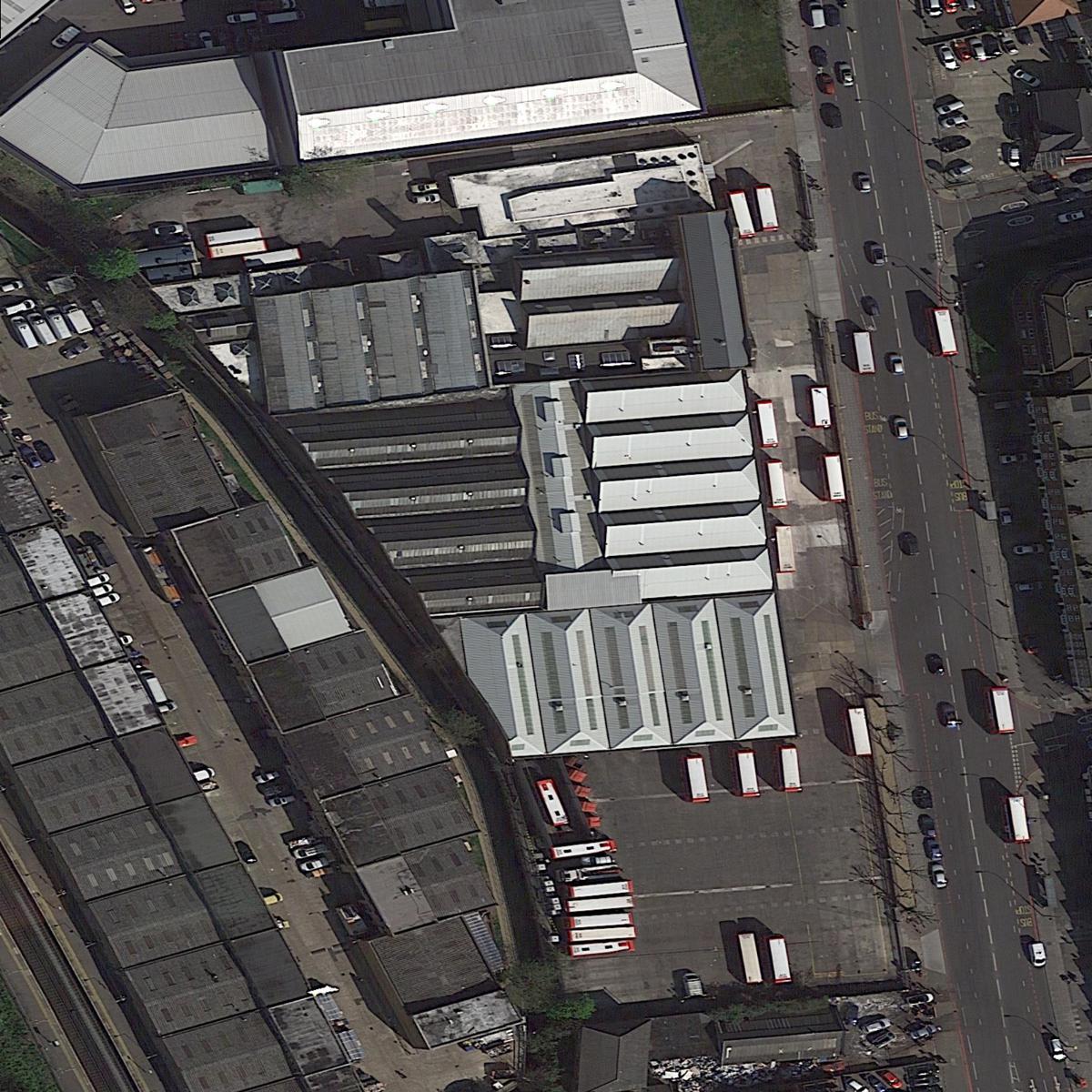}
   \caption{The original image}
\vspace{24pt}
\end{subfigure}

\begin{subfigure}[b]{1\textwidth}
   \includegraphics[width=1\linewidth]{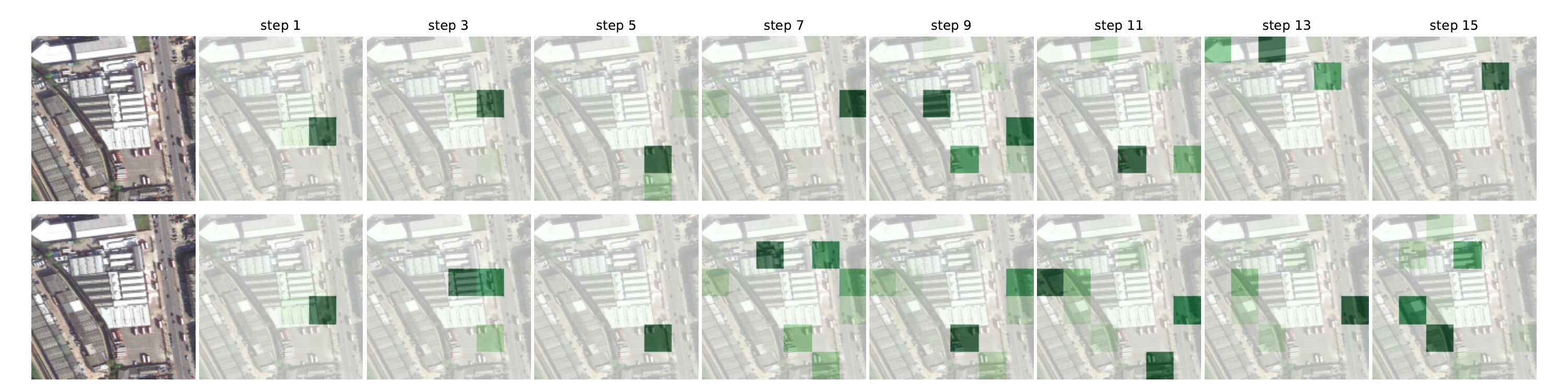}
   \caption{\textit{(\textbf{Top row})} Query sequences, and corresponding heat maps (darker indicates higher probability), obtained using VAS while enforcing the query outcomes at every stage being \enquote{\textbf{unsuccessful}}. \textit{(\textbf{Bottom row})} Query sequences, and corresponding heat maps (darker indicates higher probability), obtained using VAS while enforcing the query outcomes at every stage being \enquote{\textbf{successful}}.} 
\vspace{24pt}
\end{subfigure}
\caption[Two numerical solutions]{Sensitivity Analysis of \textit{VAS} with a sample test image and \textit{large vehicle} as target class under uniform query cost.}
\label{fig:sen7}
\end{figure*}

\begin{figure*}
\centering
\begin{subfigure}[b]{0.50\textwidth}
   \includegraphics[width=1\linewidth]{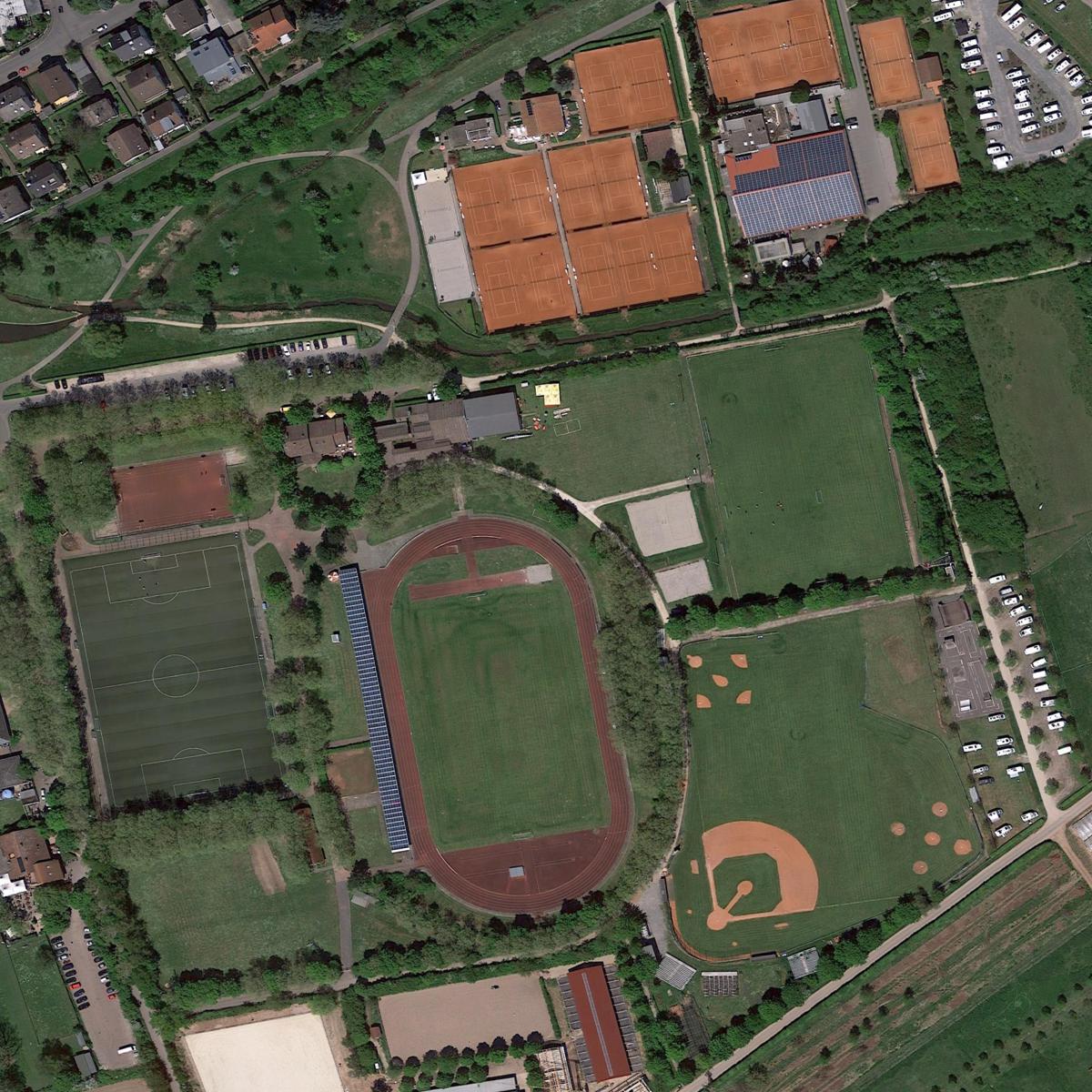}
   \caption{The original image}
\vspace{24pt}
\end{subfigure}

\begin{subfigure}[b]{1\textwidth}
   \includegraphics[width=1\linewidth]{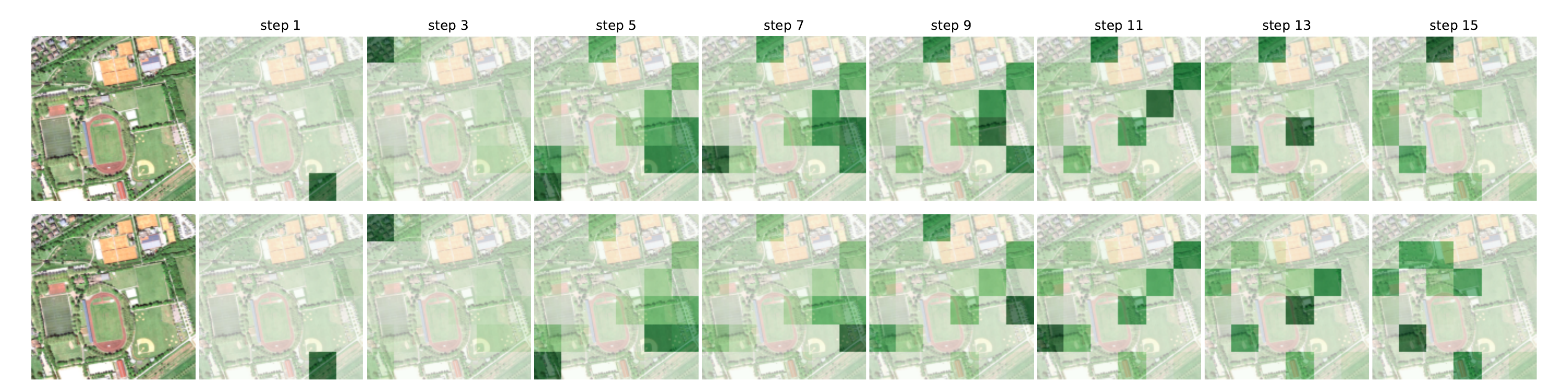}
   \caption{\textit{(\textbf{Top row})} Query sequences, and corresponding heat maps (darker indicates higher probability), obtained using VAS while enforcing the query outcomes at every stage being \enquote{\textbf{unsuccessful}}. \textit{(\textbf{Bottom row})} Query sequences, and corresponding heat maps (darker indicates higher probability), obtained using VAS while enforcing the query outcomes at every stage being \enquote{\textbf{successful}}.} 
\end{subfigure}
\vspace{24pt}
\caption[Two numerical solutions]{Sensitivity Analysis of \textit{VAS} with a sample test image and \textit{car} as target class under uniform query cost.}
\label{fig:sen8}
\end{figure*}

\begin{figure*}
\centering
\begin{subfigure}[b]{0.50\textwidth}
   \includegraphics[width=1\linewidth]{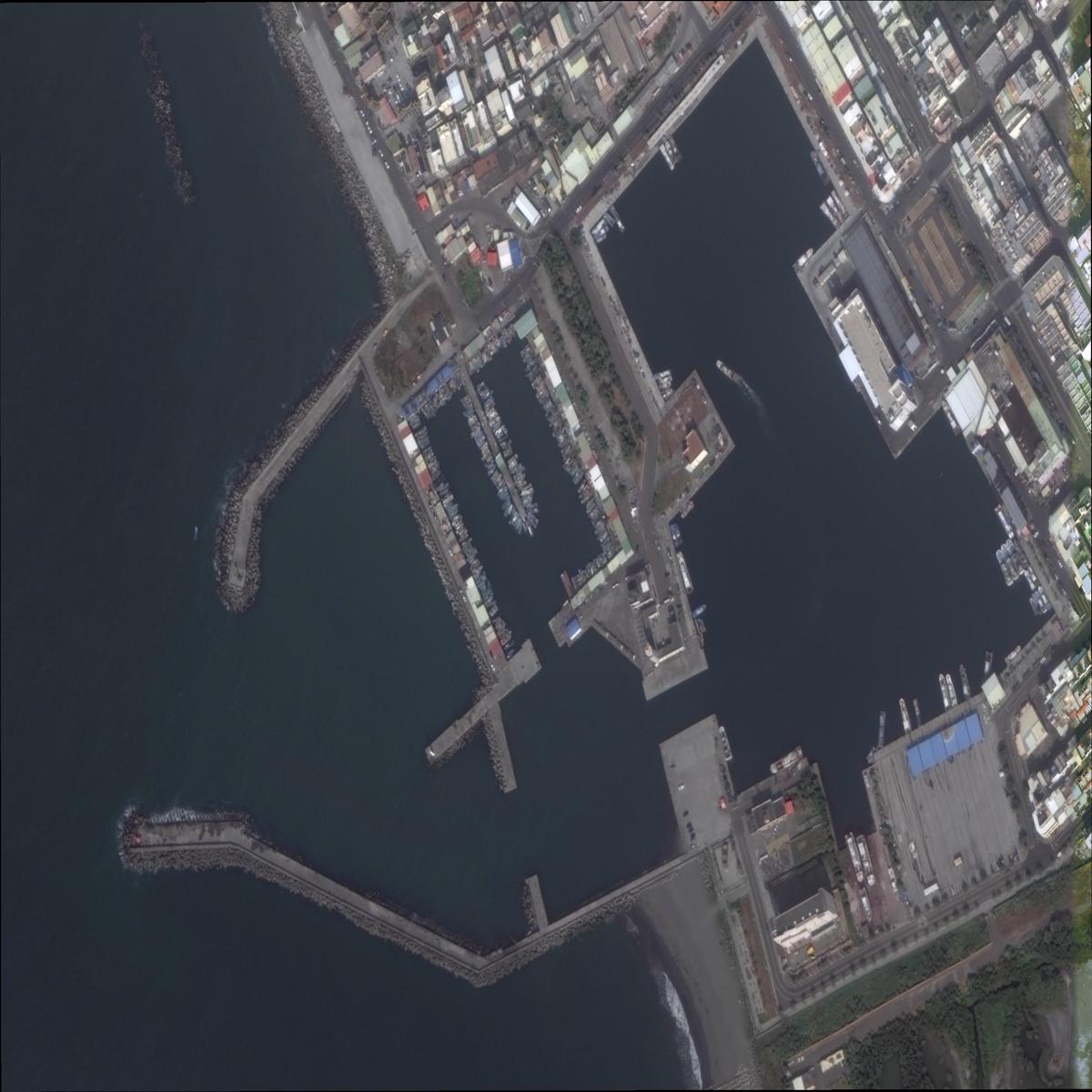}
   \caption{The original image}
\vspace{24pt}
\end{subfigure}

\begin{subfigure}[b]{1\textwidth}
   \includegraphics[width=1\linewidth]{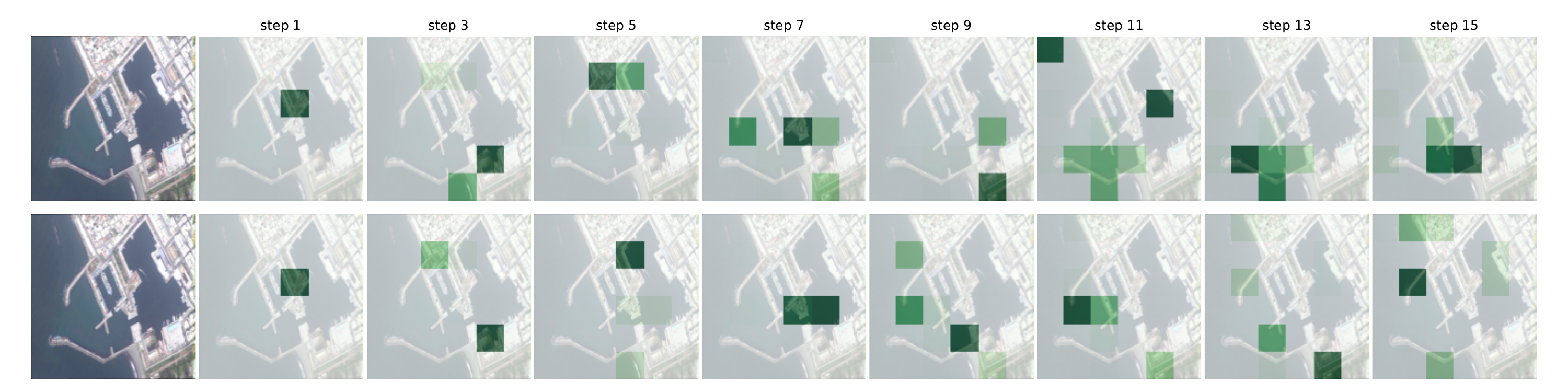}
   \caption{\textit{(\textbf{Top row})} Query sequences, and corresponding heat maps (darker indicates higher probability), obtained using VAS while enforcing the query outcomes at every stage being \enquote{\textbf{unsuccessful}}. \textit{(\textbf{Bottom row})} Query sequences, and corresponding heat maps (darker indicates higher probability), obtained using VAS while enforcing the query outcomes at every stage being \enquote{\textbf{successful}}.} 
\vspace{24pt}
\end{subfigure}

\caption[Two numerical solutions]{Sensitivity Analysis of \textit{VAS} with a sample test image and \textit{ship} as target class under uniform query cost.}
\label{fig:sen9}
\end{figure*}

We further analyze the behavior of \textit{VAS} when we intervene the outcomes of past search queries $o$ in the following ways: (i) Regardless of the \enquote{true} outcome, we set the query outcome to be \enquote{unsuccessful} at every stage of the search process and observe the change in exploration behavior of \textit{VAS}, as depicted in fig~\ref{fig:sen7}, ~\ref{fig:sen8}, ~\ref{fig:sen9}. (ii) Following a similar line, we also enforce the query outcome to be \enquote{successful} at each stage and observe how it impacts in exploration behavior of \textit{VAS}, as depicted in fig~\ref{fig:sen7}, ~\ref{fig:sen8}, ~\ref{fig:sen9}. 

\begin{figure*}
\centering
\begin{subfigure}[b]{0.50\textwidth}
   \includegraphics[width=1\linewidth]{fig_nbj/img_fig7.jpg}
   \caption{The original image}
\vspace{24pt}
\end{subfigure}

\begin{subfigure}[b]{1\textwidth}
   \includegraphics[width=1\linewidth]{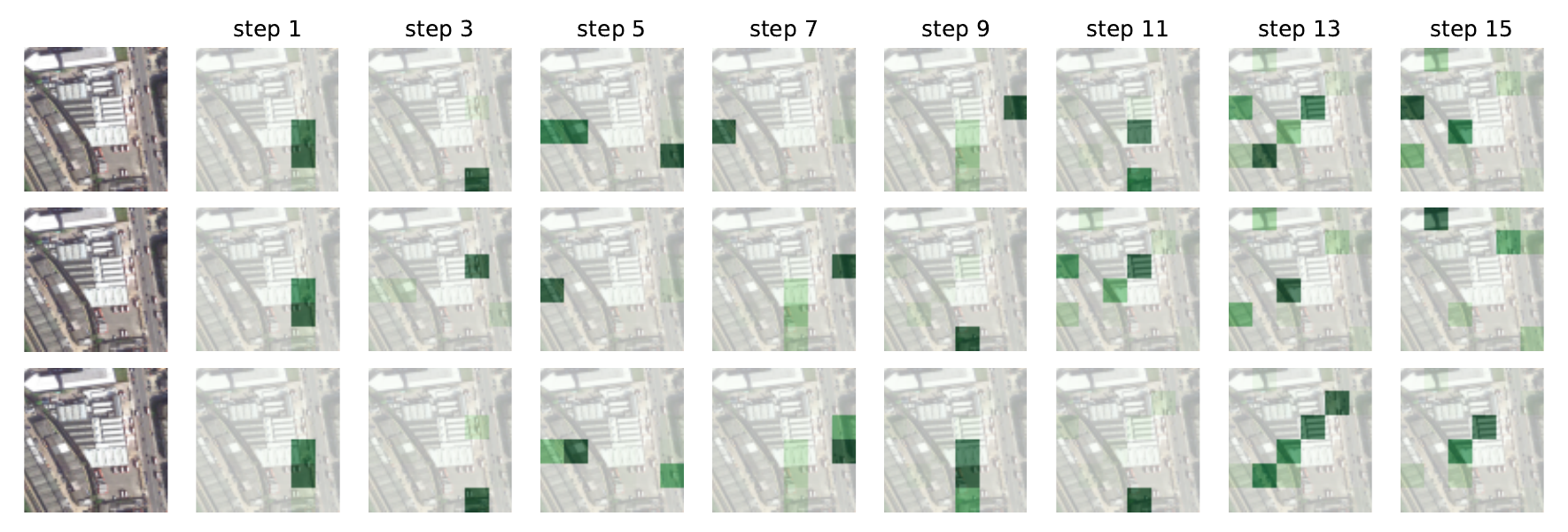}
   \caption{\textit{(\textbf{Top row})} Query sequences, and corresponding heat maps (darker indicates higher probability), obtained using VAS.
   \textit{(\textbf{Middle row})} Query sequences, and corresponding heat maps (darker indicates higher probability), obtained using VAS while enforcing the query outcomes at every stage being \enquote{\textbf{unsuccessful}}.
   \textit{(\textbf{Bottom row})} Query sequences, and corresponding heat maps (darker indicates higher probability), obtained using VAS while enforcing the query outcomes at every stage being \enquote{\textbf{successful}}.} 
\vspace{24pt}
\end{subfigure}
\caption[Two numerical solutions]{Sensitivity Analysis of \textit{VAS} with a sample test image and \textit{large vehicle} as target class under distance based query cost.}
\label{fig:senD1}
\end{figure*}

\begin{figure*}
\centering
\begin{subfigure}[b]{0.50\textwidth}
   \includegraphics[width=1\linewidth]{fig_nbj/img_fig8.jpg}
   \caption{The original image}
\vspace{24pt}
\end{subfigure}

\begin{subfigure}[b]{1\textwidth}
   \includegraphics[width=1\linewidth]{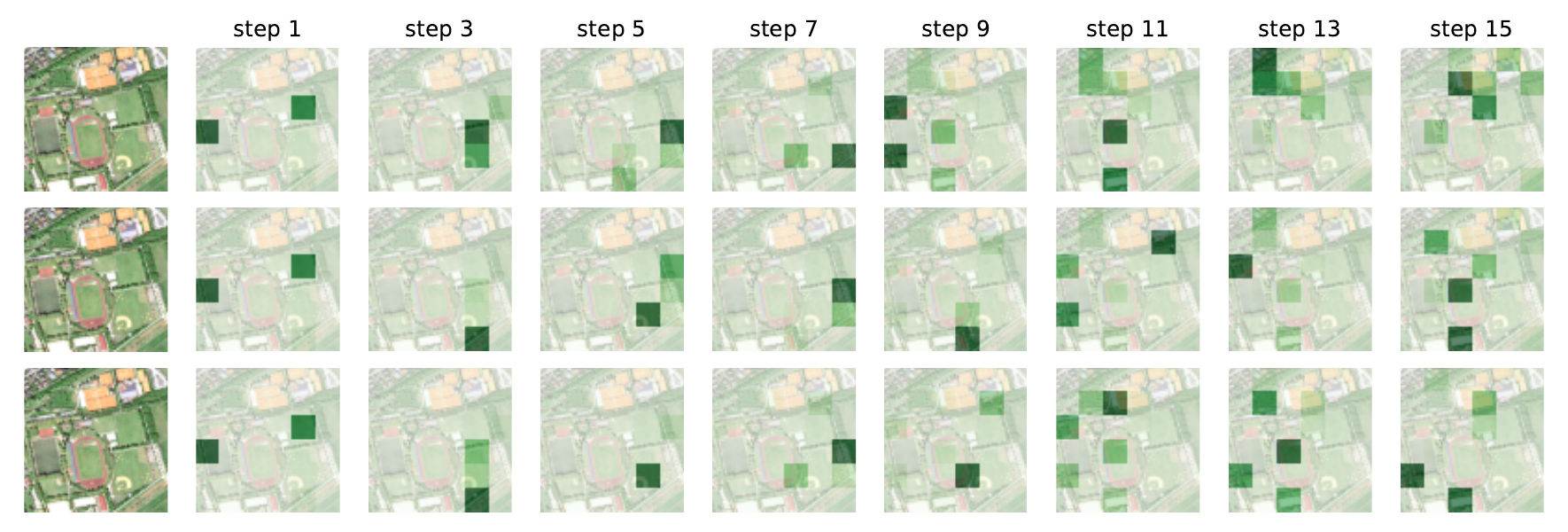}
   \caption{\textit{(\textbf{Top row})} Query sequences, and corresponding heat maps (darker indicates higher probability), obtained using VAS.
   \textit{(\textbf{Middle row})} Query sequences, and corresponding heat maps (darker indicates higher probability), obtained using VAS while enforcing the query outcomes at every stage being \enquote{\textbf{unsuccessful}}.
   \textit{(\textbf{Bottom row})} Query sequences, and corresponding heat maps (darker indicates higher probability), obtained using VAS while enforcing the query outcomes at every stage being \enquote{\textbf{successful}}.} 
\vspace{24pt}
\end{subfigure}
\caption[Two numerical solutions]{Sensitivity Analysis of \textit{VAS} with a sample test image and \textit{car} as target class under distance based query cost.}
\label{fig:senD2}
\end{figure*}

\begin{figure*}
\centering
\begin{subfigure}[b]{0.50\textwidth}
   \includegraphics[width=1\linewidth]{fig_nbj/img_fig9.jpg}
   \caption{The original image}
\vspace{24pt}
\end{subfigure}

\begin{subfigure}[b]{1\textwidth}
   \includegraphics[width=1\linewidth]{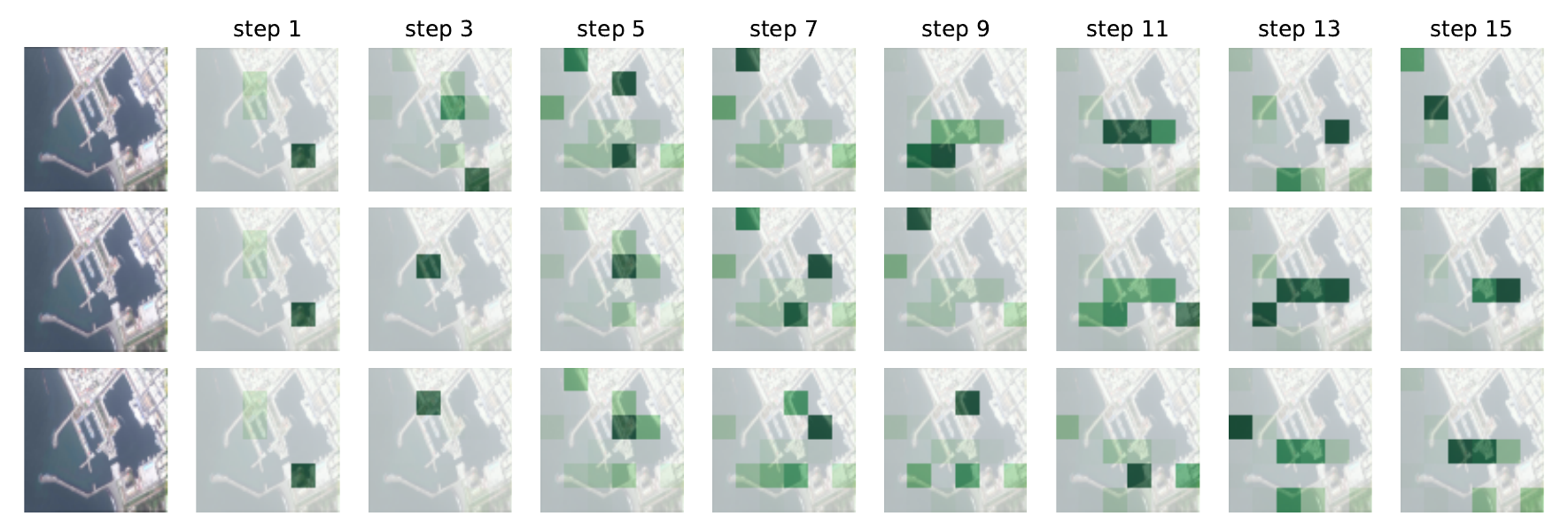}
   \caption{\textit{(\textbf{Top row})} Query sequences, and corresponding heat maps (darker indicates higher probability), obtained using VAS.
   \textit{(\textbf{Middle row})} Query sequences, and corresponding heat maps (darker indicates higher probability), obtained using VAS while enforcing the query outcomes at every stage being \enquote{\textbf{unsuccessful}}.
   \textit{(\textbf{Bottom row})} Query sequences, and corresponding heat maps (darker indicates higher probability), obtained using VAS while enforcing the query outcomes at every stage being \enquote{\textbf{successful}}.} 
\vspace{24pt}
\end{subfigure}
\caption[Two numerical solutions]{Sensitivity Analysis of \textit{VAS} with a sample test image and \textit{ship} as target class under distance based query cost.}
\label{fig:senD3}
\end{figure*}

Early VAS steps are similar between strictly positive and strictly negative feedback scenarios. This is due to the grid prediction network's input similarity in early stages of VAS. The imagery and search budget are constant between the two, and the grid state vector between the two are mostly the same (as they are both initialized to all zeros). Following from step 7 we see VAS diverge. A pattern that emerges is that when VAS receives strictly negative feedback, it begins to randomly explore. After every unsuccessful query, VAS learns that similar areas are unlikely to contain objects of interest and so it rarely visits similar areas. This is most clear in figure~\ref{fig:sen9} where we see at step 11 it explores an area that's completely water. It then visits a distinctive area that's mostly water but with land (and no harbor infrastructure). In strictly positive feedback scenarios we see VAS aggressively exploit areas that are similar to ones its already seen, as those areas have been flagged as having objects of interest. Consider the bottom row for each of figures ~\ref{fig:sen7}, ~\ref{fig:sen8}, and ~\ref{fig:sen9}. In figure ~\ref{fig:sen7}, after a burn in phase we see VAS looking at roadsides starting in step 9. In figure ~\ref{fig:sen8}, VAS seeks to capture roads. By step 15, VAS has an elevated probability for nearly the entire circular road in the upper left of the image. In figure ~\ref{fig:sen9}, VAS seeks out areas that look like harbors. Together these examples demonstrate a key feature of reinforcement learning: the ability to explore and exploit. Additionally, they show that VAS is sensitive to query results and uses the grid state to guide its search. In fig~\ref{fig:senD1}, ~\ref{fig:senD2}, ~\ref{fig:senD3}, we provide a similar visualization of VAS under Manhattan distance based query cost.

\section{Efficacy of TTA on Search Tasks involving Large Number of Grids}
We conduct experiments with number of grids $N$ as $900$. We train VAS using small car as target while evaluate with building as target class. We report the result in table~\ref{tab:table_TTA_900}. We observe a significant improvement (up to $~\textbf{4\%}$) in search performance by leveraging TTA in our proposed VAS framework. Specifically, the performance gap becomes more noticeable as the search budget increases. We observe a similar trend when we train VAS with building as target and evaluate using small car as target as presented in table~\ref{tab:table_TTA_900_1}. Such results reinforce the importance of TTA in scenarios (especially when the search budget is large) when the search target differs between training and execution environments. 

\begin{table}[h!]
       \footnotesize
        \centering
        \caption{\small{Comparative results on xView dataset with \textit{small car} and \textit{Building} as the target class during training and inference respectively under uniform query cost setting.}}
        \begin{tabular}{lcccc}
        \toprule
             \textbf{Method} &  $\mathcal{C}=18$ & $\mathcal{C}=24$ & $\mathcal{C}=30$ & $\mathcal{C}=60$  \\
            \midrule
             \emph{without TTA} ($N=900$)  & 3.32 & 4.30 & 5.41 & 10.39 \\
             \textit{\textbf{Stepwise TTA}} ($N=900$)  & 3.38 & 4.37 & 5.54 & 10.68 \\
             \textit{\textbf{Online TTA}} ($N=900$) & \textbf{3.41} & \textbf{4.42} & \textbf{5.60} & \textbf{10.81} \\
             \bottomrule
        \end{tabular}
        \vspace{-2pt}
        \label{tab:table_TTA_900}
\end{table}

\begin{table}[h!]
       \footnotesize
        \centering
        \caption{\small{Comparative results on xView dataset with \textit{building} and \textit{small car} as the target class during training and inference respectively under uniform query cost setting. }}
        \begin{tabular}{lcccc}
        \toprule
             \textbf{Method} &  $\mathcal{C}=18$ & $\mathcal{C}=24$ & $\mathcal{C}=30$ & $\mathcal{C}=60$  \\
            \midrule
             \emph{without TTA} ($N=900$)  & 1.61 & 2.07 & 2.60 & 4.93 \\
             \textit{\textbf{Stepwise TTA}} ($N=900$)  & 1.63 & 2.10 & 2.66 & 5.04 \\
             \textit{\textbf{Online TTA}} ($N=900$) & \textbf{1.66} & \textbf{2.15} & \textbf{2.71} & \textbf{5.12} \\
             \bottomrule
        \end{tabular}
        \vspace{-3pt}
        \label{tab:table_TTA_900_1}
\end{table}

\section{Saliency map visualization of \textit{VAS}}

\begin{figure*}
\centering
\begin{subfigure}[b]{0.55\textwidth}
   \includegraphics[width=0.97\linewidth]{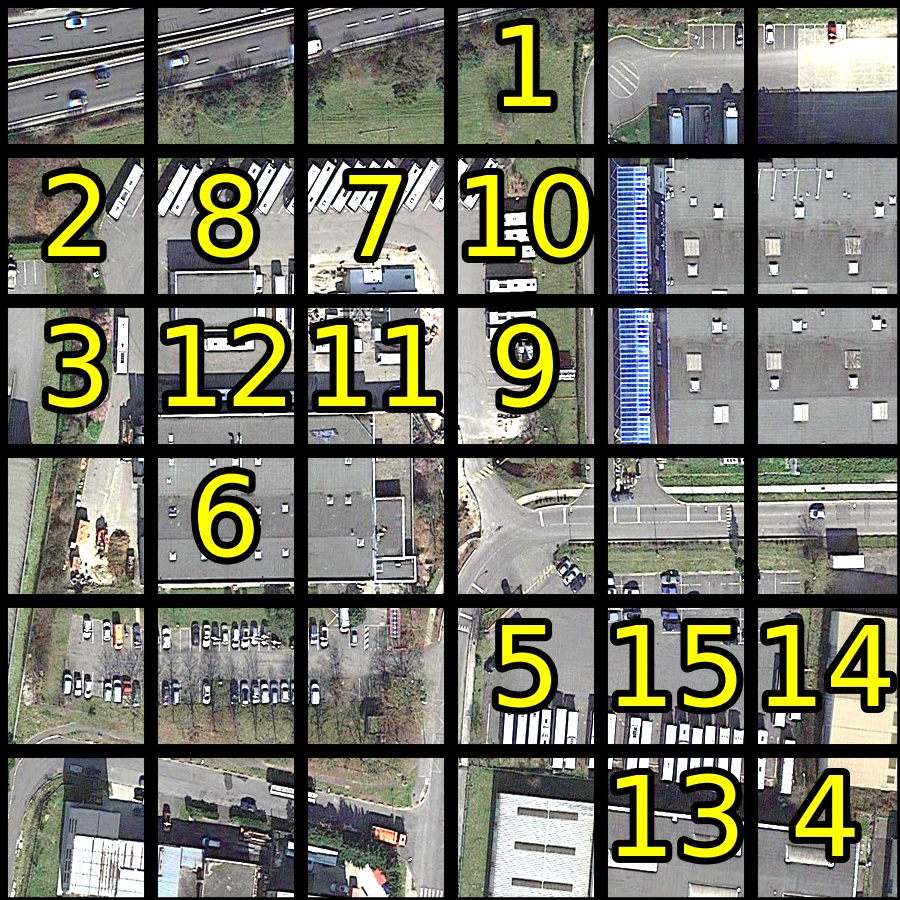}
   \caption{The original image with query sequence.}
   \label{fig:1} 
\vspace{30pt}
\end{subfigure}
\begin{subfigure}[b]{0.97\textwidth}
   \begin{tabular}{cccc}
       step 1 & step 5 & step 10 & step 15 \\
       \includegraphics[width=0.23\linewidth]{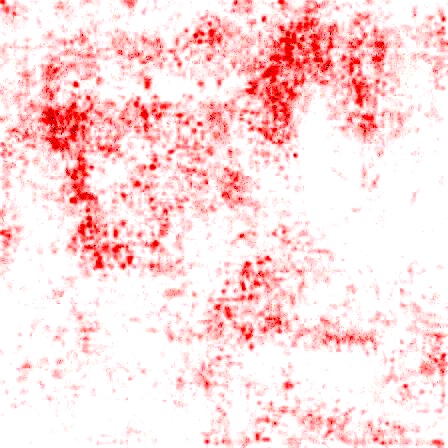} &
       \includegraphics[width=0.23\linewidth]{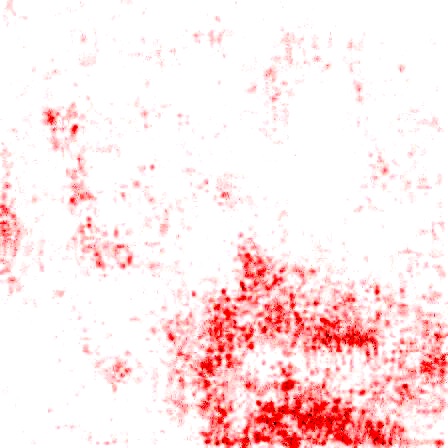} &
       \includegraphics[width=0.23\linewidth]{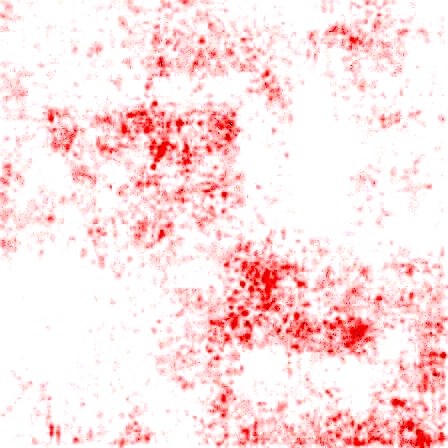} &
       \includegraphics[width=0.23\linewidth]{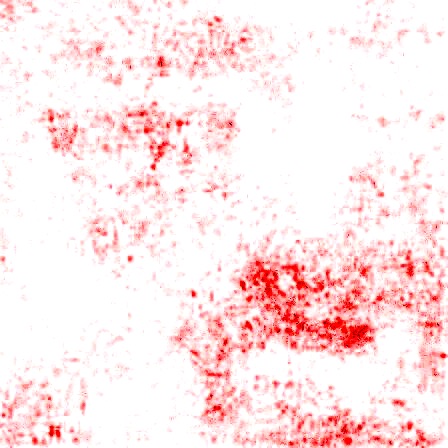} \\
   \end{tabular}  
   \caption{Saliency maps (red indicates high saliency), obtained using VAS at different stages of search process with \textit{large vehicle} as target.}
\end{subfigure}

\caption[Two numerical solutions]{Saliency map visualization of \textit{VAS} under uniform cost budget.}
\label{fig:sal1}
\end{figure*}

\begin{figure*}
\centering
\begin{subfigure}[b]{0.55\textwidth}
   \includegraphics[width=0.97\linewidth]{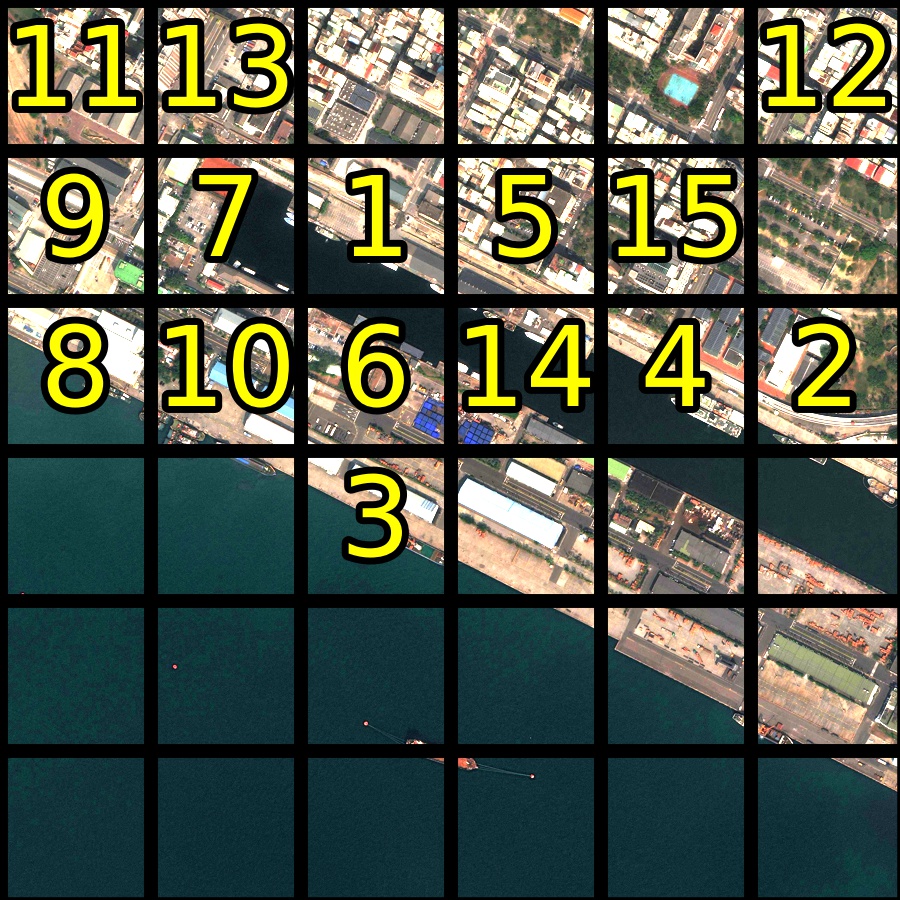}
   \caption{The original image with query sequence.}
   \label{fig:2} 
\vspace{30pt}
\end{subfigure}

\begin{subfigure}[b]{0.97\textwidth}
   \begin{tabular}{cccc}
       step 1 & step 5 & step 10 & step 15 \\
       \includegraphics[width=0.23\linewidth]{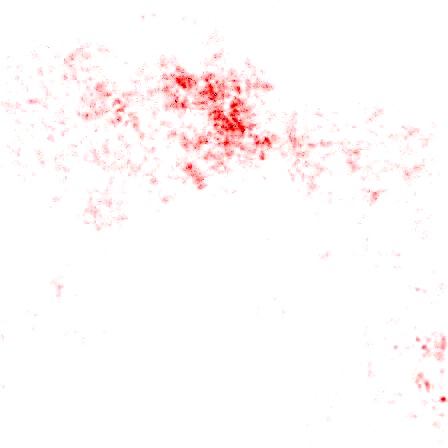} &
       \includegraphics[width=0.23\linewidth]{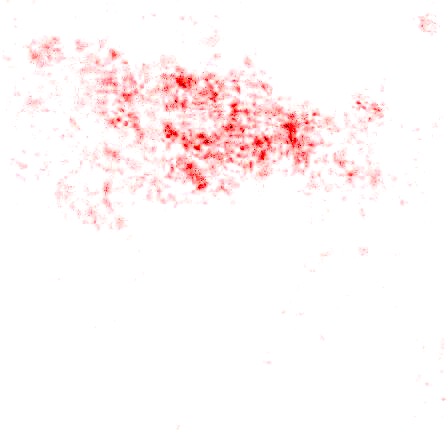} &
       \includegraphics[width=0.23\linewidth]{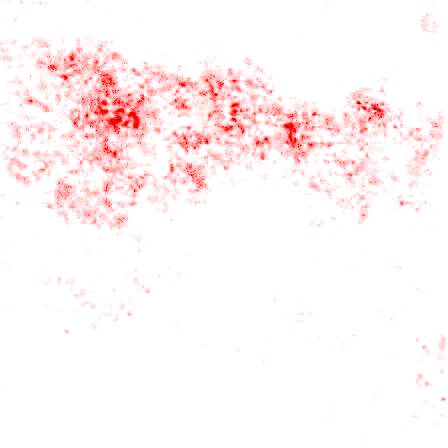} &
       \includegraphics[width=0.23\linewidth]{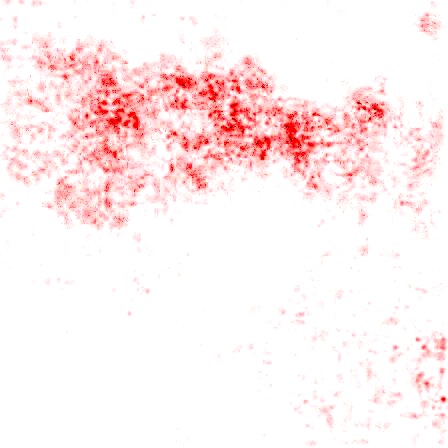} \\
   \end{tabular}  
   \caption{Saliency maps (red indicates high saliency), obtained using VAS at different stages of search process with \textit{small car} as target.}
\end{subfigure}

\caption[Two numerical solutions]{Saliency map visualization of \textit{VAS} under uniform cost budget.}
\label{fig:sal2}
\end{figure*}

\begin{figure*}
\centering
\begin{subfigure}[b]{0.55\textwidth}
   \includegraphics[width=0.97\linewidth]{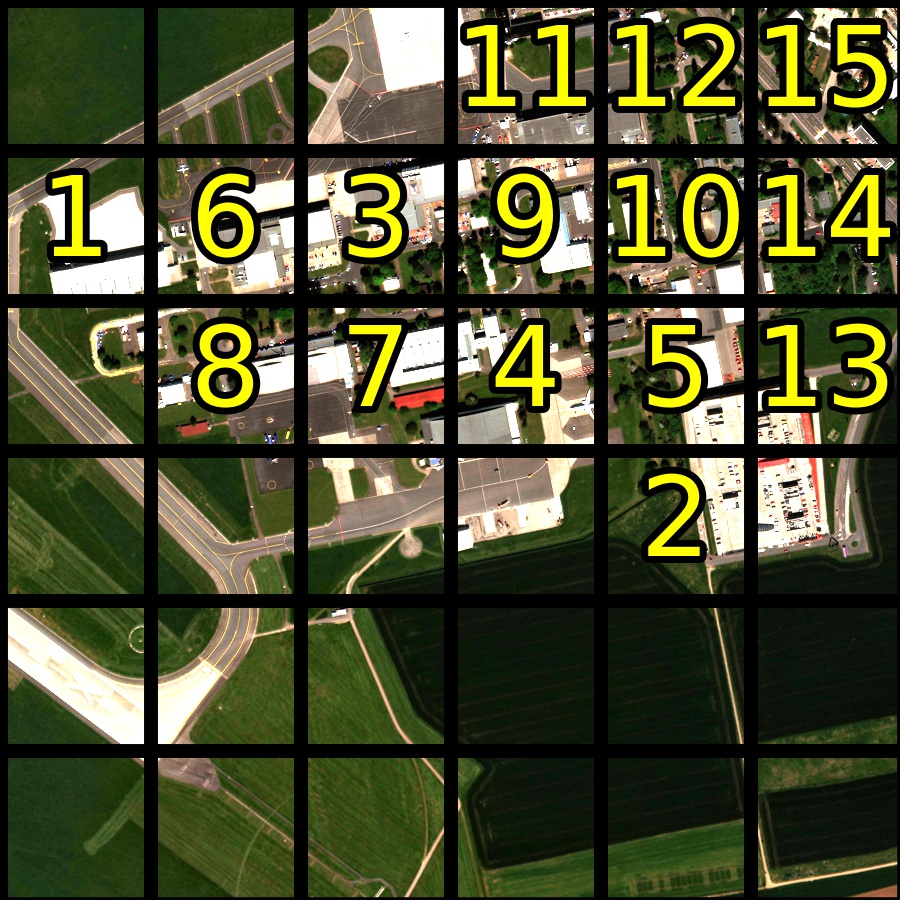}
   \caption{The original image with query sequence.}
   \label{fig:3} 
\vspace{30pt}
\end{subfigure}

\begin{subfigure}[b]{0.97\textwidth}
   \begin{tabular}{cccc}
       step 1 & step 5 & step 10 & step 15 \\
       \includegraphics[width=0.23\linewidth]{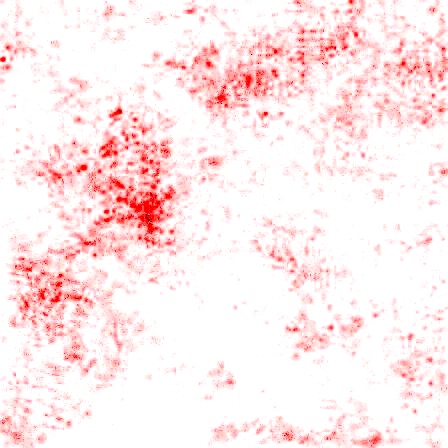} &
       \includegraphics[width=0.23\linewidth]{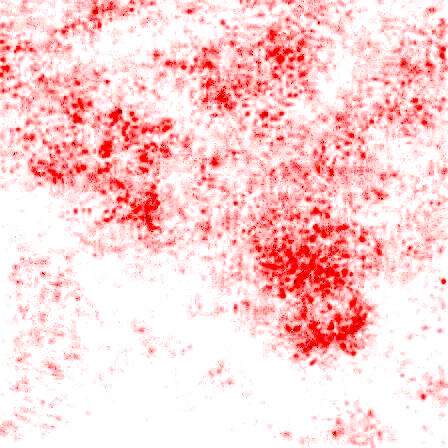} &
       \includegraphics[width=0.23\linewidth]{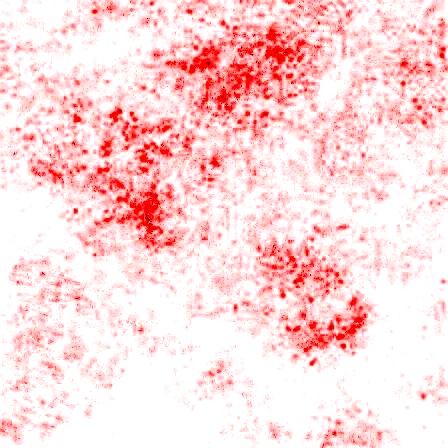} &
       \includegraphics[width=0.23\linewidth]{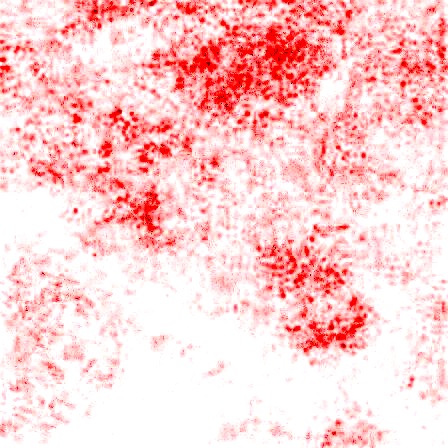} \\
   \end{tabular}  
   \caption{Saliency maps (red indicates high saliency), obtained using VAS at different stages of search process with \textit{small car} as target.}
\end{subfigure}

\caption[Two numerical solutions]{Saliency map visualization of \textit{VAS} under uniform cost budget.}
\label{fig:sal3}
\end{figure*}

In Figure (\ref{fig:sal1},\ref{fig:sal2},\ref{fig:sal3}), we show the saliency maps obtained using a pre-trained VAS policy at different stages of the search process. Note that, at every step, we obtain the saliency map by computing the gradient of the output that corresponds to the query index with respect to the input. Figure~\ref{fig:sal1} corresponds to the large vehicle target class while the Figure~\ref{fig:sal2} and Figure~\ref{fig:sal3} correspond to the small vehicle. All saliency maps were obtained using the same search budget (K = 15). These visualizations capture different aspects of the VAS policy. Figure~\ref{fig:sal1} shows its adaptability, as we see how heat transfers from non-target grids to the grids containing targets as search progresses. By comparing saliency maps at different stages of the search process, we see that, VAS explores different regions of the image at different stages of search, illustrating that our approach implicitly trades off exploration and exploitation in different ways as search progresses. Figure~\ref{fig:sal2} shows the effect of supervised training on VAS policy. If we observe the saliency maps across time, we see that VAS never searches for small vehicles in the sea, having learned not to do this from training with similar images. Additionally, we notice that the saliency map’s heat expands from left to right as the time step increases, encompassing more target grids, leading to the discovery of more target objects. We observe similar phenomena in figure~\ref{fig:sal3}. We can see that while earlier in the search process queries tend to be less successful, as the search evolves, our approach successfully identifies a cluster of grids that contain the desired object, exploiting spatial correlation among them. Additionally, at different stages of the search process, VAS identifies different clusters of grids that include the target object.

